WILEY

# Deep learning-based crop row detection for infield navigation of agri-robots

Rajitha de Silva[1] | Grzegorz Cielniak[1] | Gang Wang[2] | Junfeng Gao[1]

[1]Lincoln Agri-Robotics Centre, Lincoln Institute for Agri-Food Technology, University of Lincoln, Lincoln, UK

[2]Center of Brain Sciences, Beijing Institute of Basic Medical Sciences, Beijing, China

**Correspondence**
Gang Wang, Center of Brain Sciences, Beijing Institute of Basic Medical Sciences, 100850 Beijing, China.
Email: g_wang@foxmail.com

Junfeng Gao, Lincoln Agri-Robotics Centre, Lincoln Institute for Agri-Food Technology, University of Lincoln, Lincoln, UK.
Email: jugao@lincoln.ac.uk

**Funding information**
Lincoln Agri-Robotics, Excellence in England (E3) Programme; Beijing Municipal Natural Science Foundation, Grant/Award Number: 4214060; National Natural Science Foundation of China, Grant/Award Number: 62102443

**Abstract**

Autonomous navigation in agricultural environments is challenged by varying field conditions that arise in arable fields. State-of-the-art solutions for autonomous navigation in such environments require expensive hardware, such as Real-Time Kinematic Global Navigation Satellite System. This paper presents a robust crop row detection algorithm that withstands such field variations using inexpensive cameras. Existing data sets for crop row detection do not represent all the possible field variations. A data set of sugar beet images was created representing 11 field variations comprised of multiple grow stages, light levels, varying weed densities, curved crop rows, and discontinuous crop rows. The proposed pipeline segments the crop rows using a deep learning-based method and employs the predicted segmentation mask for extraction of the central crop using a novel central crop row selection algorithm. The novel crop row detection algorithm was tested for crop row detection performance and the capability of visual servoing along a crop row. The visual servoing-based navigation was tested on a realistic simulation scenario with the real ground and plant textures. Our algorithm demonstrated robust vision-based crop row detection in challenging field conditions outperforming the baseline.

## 1 | INTRODUCTION

The increase in demand for food production leads to increased labor for agricultural tasks. Agricultural robots are a crucial need to meet this increasing demand for labor in agricultural environments. Uncertainty of performance in agricultural technology has been identified as a key concern with the adopters of new technologies (Gallardo & Sauer, 2018). Navigation in row crop fields poses uncertainties, such as weed density, varying light levels, growth stages, and discontinuities of crop row. Such uncertainty could be mitigated to a certain extent by using an accurate Global Navigation Satellite System (GNSS) system to aid the navigation (Bengochea-Guevara et al., 2016). Such accurate GNSS systems are expensive and they will not always provide coverage for all kinds of environments, for example, in hilly terrains. Computer vision algorithms are important components that need to be improved to promote the current agricultural systems (Oliveira et al., 2021). Computer visions sensors are often cheaper than Real-Time Kinematic Global Positioning System (RTK-GPS) and other expensive navigation hardware being used in agricultural robots. However, the technological hurdles pertaining to challenging field

---

conditions and the uncertainty of using computer vision in agricultural environments must be addressed to reap the benefit of such cheap hardware in agricultural robots (Wang et al., 2022).

Earlier work on crop row detection was based on color-based segmentation approaches (Bonadies & Gadsden, 2019). These approaches are generally based on a color-based segmentation method to extract green color regions of an image corresponding to crop plants followed by some filtering approaches to remove noise from the segmented image. The Hough transform (Ballard, 1981) is used on this segmented mask to detect the lines present as crop rows (Ji & Qi, 2011). Existing work in this domain presents case-specific solutions which would only work in a given type of crop under a given set of environmental conditions.

Deep learning is a branch of machine learning that is often used in object detection and segmentation of image data. Some of these limitations in the traditional approaches have been proven to be overcome with the novel deep learning-based approaches (Pang et al., 2020). Deep learning algorithms are capable of identifying patterns in images despite the variations of light levels and minor occlusions. To this end, deep learning algorithms could detect crop rows despite the variations in light levels, size and placement of the objects, occlusions due to weeds, and variations in the background of the desired subject (crop rows). These deep learning algorithms are capable of detecting high-level semantic features such as crop rows rather than just isolating the pixels belonging to the crop. The work in deep learning-based crop row detection is mostly realized by image segmentation approaches to predict a crop row mask as a binary image. The predicted mask is then used to detect the parameters defining lines corresponding to crop rows. The data sets used in existing work often represent one or a few field variations (Ahmadi et al., 2022; Bah et al., 2019; Vidović et al., 2016). A comprehensive data set must be used to train these algorithms in order for them to predict crop rows in challenging field conditions.

Crop row detection is the initial step toward autonomous navigation in crop rows. A selection procedure is required to identify the primary crop row which must be followed by the robot among the lines detected in the crop row mask. The angular difference between the primary crop row and the robot heading direction needs to be calculated to generate the steering command to control the robot (Adhikari et al., 2020; Ahmadi et al., 2020). While the deep learning algorithms are capable of generating a reliable crop row mask based on input images, a robust algorithm is needed to identify the central crop row which the robot intends to follow.

A robot often encounters uneven terrain during the navigation in an arable field. Such uneven terrain causes the robot to drift and change its heading direction in a real environment. The evaluation of the effect of crop row detection algorithm on crop row navigation was done in a controlled simulation environment to avoid external disturbances, such as uneven terrain.

The main contributions of this work are summarized as follows:

- A unique data set of sugar beet crop row images collected during multiple crop seasons representing diverse growth stages, light levels, varying weed densities, curved crop rows, and discontinuities.
- A novel crop row selection algorithm which accurately predicts central crop row in any field condition including growth stage, weed density or discontinuities without needing condition-specific tuning, using a crop row segmentation mask.
- Evaluation of crop row detection performance under varying field conditions using a combined performance indicator which accounts for both angular and positional errors of the detected crop row.
- A visual servoing-based crop row navigation algorithm tested on a realistic simulation to evaluate the visual servoing performance of our method.

The remainder of this paper is arranged as follows: Section 2 discusses the existing approaches and their shortcomings. Section 3 introduces the data set. Section 4 explains the methods followed in training the deep learning-based crop row detection algorithm and visual servoing controller. Section 5 summarizes the results. Section 6 outlines the possible future developments. Section 7 concludes the outcomes of the proposed research.

## 2 | RELATED WORK

### 2.1 | Classic computer vision-based methods

Crop row detection is an integral part of the open-field navigation of agricultural robots. Contour tracing or edge detection is also a popular method of crop row detection despite the availability of color segmentation-based approaches, such as Ahmadi et al. (2020) and Xue et al. (2012). The contour detection approach requires adequate image filtering for noise reduction and adjusting the algorithm thresholds to adapt to the varying lighting conditions (Bonadies & Gadsden, 2019). Although this approach is viable under known lighting conditions, further research must be conducted on the dynamic threshold setting under varying ambient lighting. Gao et al. (2018) also discussed the use of the Hough transform (Ballard, 1981) in crop row detection. The Hough transform-based crop row detection only limits to straight crop rows whereas crop rows with a curvature will be approximated to a straight line. This limitation could be mitigated at the cost of limiting the row detection region of the image when performing the Hough transform. However, executing Hough transform on a realtime frame-by-frame basis is a highly computationally intensive task. Hence a robot with Hough transform-based crop row detection needs an expensive and powerful embedded computation unit onboard (Bah et al., 2019; Winterhalter et al., 2018). In addition to the above considerations, the speed of the robot, speed of image acquisition, computational time, and delays of execution in motor control must be accounted for when developing the autonomous navigation algorithms for agricultural robots.

Ahmadi et al. (2020) presented a visual navigation framework for row crop fields. The system only uses onboard cameras for navigation without performing explicit localization. The switching from one raw to another is also implemented in the same framework. Researchers have identified that the agricultural environments contain lack of reliable sensor measurements to detect distinguishable landmarks, which causes high visual aliasing. They also state that the constantly growing crops make it difficult to maintain a fixed map of the environment. Their robot could: autonomously navigate through row crop fields without maintaining any global reference maps, monitor crops in the fields with high coverage by accurately following the crop rows, and the system is robust to fields with various row structures and characteristics. Crop row detection was based on the Excess Green Index calculation. Their crop row detection scenario is based on a field setup which does not have any weeds. The robotic platform is equipped with two cameras (front and back) to perform the crop transition. Their work indicates that the vision-based navigation in crop rows could be performed without the need for a global map.

Winterhalter et al. (2021) used existing crop row detection algorithms based on Hough transform and random sample consensus algorithm along with a GNSS-referenced map to navigate in crop rows. They highlight the need for GNSS due to the lack of information when switching the robot from one crop row to another. They used end-of-field detection to identify the end of traversed crop line to decide on switching the row. However, the addition of GNSS hardware rises the cost of the robot. However, they highlight the need of end of row detection in crop row detection algorithms to develop a navigation algorithm that can traverse an entire field.

Ahmadi et al. (2022) presented a crop row detection algorithm based on color-based segmentation along with a visual servoing controller. They use the Excess Green Index to create the segmentation mask and detect individual plants from the mask. The detected plant centers are used to detect crop rows with least squares fitting. They make use of the arrangement of the crop rows to identify the central crop row. We will use this as our baseline to compare our method since it is sufficiently novel and the application scenario is similar to our methods and data set. The extended reasoning behind the selection of this baseline is explained in Section 5.1.

## 2.2 | Deep learning-based methods

Deep learning-based methods are already used to detect lanes for autonomous navigation on roads (Zhao et al., 2020). Such methods are already ahead of classical computer vision in accurately predicting lane structure under varying conditions. A deep learning-based crop row detection method was implemented in paddy fields by Adhikari et al. (2020). Their method can accurately predict crop rows under varying field conditions, such as shadows, growth stages, and row spacing. They use "mean pixel deviation" as a metric to evaluate their crop row detection performance. Their method is also capable of predicting crop rows with a varying field of view (FOV) of the robot. However, their data set is limited to 350 images and the images in the data set do not contain highly weed-populated examples or curved crop rows. Data sets with limited field conditions would not be able to train robust machine learning models that can be deployed under real-world conditions. Therefore, the generation of a data set with a wide variety of field conditions is essential for robust crop row detection systems. Lin and Chen (2019) developed a convolutional neural network (CNN)-based crop row navigation system for tea plantations. However, they rely on Hough transform to extract the edge line parameters from the segmentation mask. Although the Hough transform is a reliable method to extract line parameters in a given setting, the parameters of the Hough transform-based line extraction method must be retuned to multiple crops and different grow stages. Their method is also limited to tea plantations where tea bushes are arranged in a densely populated wide row and the CNN is trained with wider ground truth labels representing the entirety of the tea bushes. Such representations may bias the CNN to learn crop-specific features minimizing its ability to use any crop without retraining.

Autonomous crop row guidance using adaptive multi-region of interest (multi-ROI) was presented in Ponnambalam et al. (2020). The authors were addressing the field variations caused by the size of the crops through different growth stages. The crop row feature points were fitted using horizontal strips from the crop row mask generated by a CNN. Their algorithm can select adaptive ROIs for feature point selection in each horizontal strip. The image labels used in our approach represent every crop row as a size-invariant high-level semantic feature and hence they are independent of the grow stages. Doha et al. (2021) trained U-Net CNN (Ronneberger et al., 2015) with an augmented version of CRBD data set (Vidović et al., 2016). The augmentations were obtained by adding artificial objects to the images, color adjustments, and rotations. Their work suggests that the CNNs could be used to segment crop rows accurately despite the presence of occlusions and varying light levels. However, their data set does not include realistic field variations to verify the capability of their claims under real field scenarios. Bakken et al. (2021) used automatic label generation using a predefined crop row structuring model to create a crop row detection data set. Crop row center points collected using GNSS data were used to align the ground truth label with the images. They used Segnet (Badrinarayanan et al., 2017) implementation to detect crop rows with the automatically generated noisy data. They claim that the resulting crop row segmentation from the CNN was better than the noisy data set it was trained on. Although they acknowledge the presence of field variations such as different leaf colors, offshoots, and shadows, their data set does not capture all of those variations.

Bah et al. (2019) combined the predictions from two CNNs architectures to estimate accurate crop rows from unmanned aerial vehicle images. This multi-CNN architecture enables them to generate robust predictions against weeds and discontinuities. While their method does not cover all the possible field variations, it proves the capability of deep learning models to make robust predictions under varying field conditions. The data set they use could not be directly used to train crop row detection models for ground robots. They use "Intersection over Union" (IoU) of image pixels to evaluate the performance of crop row detection. Therefore this method will need additional RTK-GPS hardware to assist the navigation of a ground robot.

The CNNs are also used in end-to-end crop row following to predict control commands directly from image input (Bakken et al., 2019). The CNN was only tested in polytunnel environment. The performance of such systems under varying field conditions and open-field scenarios is yet to be tested. However, recent work on end-to-end solutions for autonomous robot navigation indicates the capability of this technology to be deployed in diverse settings (Kahn et al., 2021; Manderson et al., 2020; Pan et al., 2020). The literature does not provide sufficient evidence to confirm the potential of such systems to generalize across diverse crops and field conditions despite most of these studies indicate promising results in each of the implemented solutions (Guastella & Muscato, 2020).

The existing work on crop row detection indicates the weaknesses in classical computer vision techniques to accurately predict crop rows under varying field conditions. The RTK-GPS-based methods are accurate yet expensive and not applicable to all environments. State-of-the-art camera sensors pose a promising potential to be used in agricultural environments while being relatively cheaper. The work done on deep learning-based crop row detection has proven to generate robust crop row predictions under varying field conditions. However, the lack of comprehensive data sets has limited the researchers to develop a model that can make predictions against most of the field variations in agricultural environments.

## 2.3 | Visual servoing

Visual servo control is the use of computer vision data to control the motion of a robot using a camera mounted on the robot or in the workspace (Chaumette et al., 2016). Visual servoing schemes differ in the way the control signal is designed, including using features directly available in the image (Image-Based Visual Servoing [IBVS] control) or estimating the pose of the robot from image measurements (pose-based visual servo control). IBVS is used to control robots in diverse settings including agricultural robotics (Ahmadi et al., 2020; Cherubini et al., 2008).

The IBVS controller is modeled based on the assumption that the relative positions of the image feature points are known in the absence of perturbations. However, the perturbations could cause the IBVS controller to cause system instability due to unmodelled dynamics (Barbosa et al., 2021). Authors have added additional robustness terms to the IBVS controller to cope with instability caused by external disturbances. Bengochea-Guevara et al. (2016) implemented a visual servoing controller based on fuzzy logic for in-row navigation in open fields. They use central crop row extracted via classical computer vision-based methods to perform visual servoing.

## 2.4 | Hardware for navigation

Bakker et al. (2011) claimed to have suggested the first approach toward GNSS and vision-based crop row mapping. They mainly use a Real-Time Kinematic Differential Global Positioning System (RTK-DGPS) for odometry. A camera mounted on the robot is used to detect the crop rows during navigation and control the robot based on linear and angular errors with feedback control. The proposed system was implemented in a sugar beet field with headland areas on either side of the field for the robot to make U-turns to resume navigation in the adjacent line. A front-mounted camera is used for crop raw detection. The presented results indicate a standard deviation of the lateral error at 1.6 cm and a heading error at 0.008 rad. However, the error readings at two adjacent rows indicate that the error is slightly increased after the robot navigates the headland path to re-enter the next row. Although the authors assume this was due to roll caused by the bumpier headland, provided results are insufficient to justify this claim.

Tracking cameras such as ZED mini and Intel Realsense T265 are considered a vision-based replacement for GNSS-based odometry (Mahmoud & Atia, 2019). Agarwal et al. (2020) conducted an independent evaluation of the Intel Realsense Tracking Camera (T265) for an Unmanned Aerial System (UAS) use case. T265 is an off-the-shelf tracking camera which provides visual SLAM (V-SLAM) with two fish eye lens sensors (163 ± 5° FOV), BMI055 inertial measurement unit (IMU), and a built-in vision processing unit (VPU). The T265 camera processes the sensor data with Intel Movidius Myraid 2 VPU and the algorithms used in the devices is closed source. The raw and Extended Kalman Filter-Based estimation and yaw angle data from the T265 are compared with the Vicon motion capture system for ground truth. Multiple tests were conducted using UAS flight missions each of which lasted between 20 and 30 s. The initial tests demonstrated a 20-cm maximum mean position error while it was reduced to a 10-cm maximum mean position error during the latter tests. The researchers believe that the improvement of accuracy over the tests is a result of caching detected feature points without purging them upon reinitialization. However, the capacity or the frequency of such cache is unknown. The mean heading error of the device was 3°. Although this study is solely based on the data generated by the T265 tracking camera, several other studies (Bayer & Faigl, 2019; Mahmoud & Atia, 2019) and the reference material of Intel Realsense indicate that the accuracy of T265-based navigation systems could be improved by combined usage of other hardware, such as depth cameras or other odometry sources.

The importance of the FOV of the cameras used in vision-based navigation systems was indicated by Xue et al. (2012) in earlier work on crop row navigation. The researchers used a single monocular camera with a variable FOV setup for better accuracy in navigation at the end of crop rows. Thus the researchers identify the requirement of a wider FOV toward the success of vision-based crop row navigation.

## 3 | DATA SET

We have created a crop row data set in a sugar beet field from images captured from a front-mounted camera in a Husky robot. The data set contains images representing 11 field variations and 43 possible combinations of those field variations. The data set presented in this paper is the succession of Crop Row Detection Lincoln Data set (CRDLD) (de Silva et al., 2022). The CRDLD version 1 only had 10 data categories where each category represents a single field variation. The CRDLD version 2 presented in this paper examines 11 categories of field variations along with the combinations of those 11 categories. CRDLDv2 has been added with additional camera views, RTK-GPS data, and velocity commands given to the robot while driving along the crop rows. We are only using the RGB images from the front-mounted camera for our work in this paper.

## 3.1 | Hardware setup

A forward-facing camera pair: Intel Realsense D435i RGB-D camera and T265 tracking camera were equipped at a pitch angle of −25° to collect forward-looking pictures of crop rows along with visual odometry. The depth frame had a FOV of 87° × 58° (Horizontal × Vertical) and the RGB frame had a FOV of 69° × 42° (Horizontal × Vertical). Two additional D435i RGB-D cameras were mounted on either side of the robot pointing toward the adjacent crop rows as shown in Figure 1. An EMLID Reach RS-Plus RTK-GPS was mounted on the robot with Network Transport of RTCM via Internet Protocol corrections. The GNSS readings reported an average accuracy of ±4cm. The robot was straddling the crop row while it has been driven along the crop row. The velocity commands to keep the robot in line were provided by a human using a Bluetooth controller. Although the crop rows were straight, the rocks present in the soil caused the robot to require constant steering to follow the line in the forward direction. A summary of all the raw data collected using the sensor array is given in Table 1. The RGB-D images captured by the forward-facing D435i camera were mainly used in the work presented in this paper. The remaining data will be used for the future work of this project in developing a vision-based navigation system for crop rows.

## 3.2 | Data categories

The CRDLDv2 encompasses crop row images under 11 main categories as listed in Table 2. These 11 categories account for various field variations, including shadows, growth stages, weed densities, and light levels. Table 3 indicates the 43 data classes derived from the 66 possible combinations of these 11 data categories. These 43 combinations are chosen based on their relevance to observing the field variations and their effect on crop row detection. The combinations which have been ignored (gray-colored cells) are either improbable to occur or unable to capture due to adverse weather during the data collection period. Examples for each of the data classes listed in Table 3 are given in Figure 2. Each class contains 25 labeled RGB-D images totaling 1125 images in total in the data set. (The data set can be accessed with the following link: https://github.com/JunfengGaolab/CropRowDetection). The images for each class are mostly drawn from the same crop row while some classes have images drawn from a few different crop rows. Sample images from each class are shown in Figure 3. While all images in each class share two given unique field conditions, they have one or many other field conditions present in them. For example, class 43 consists of 25 unique images which have tire tracks and discontinuities present in all of them. However, these images have varying weed densities, cloudy or sunny weather, and straight or curved crop rows.

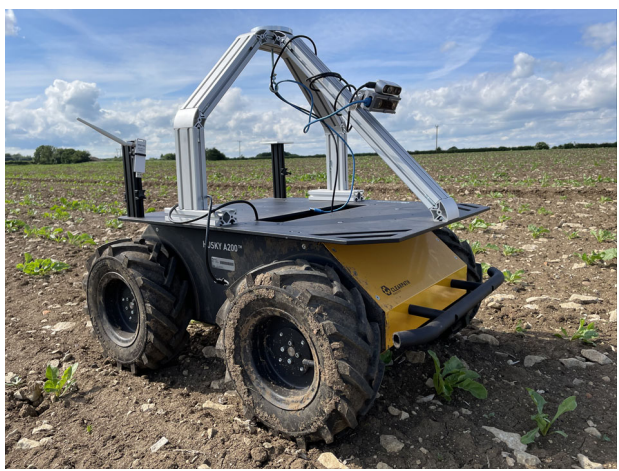

**FIGURE 1** Husky robot with Realsense cameras D435i (top) and T265 (bottom) in a sugar beet field. [Color figure can be viewed at wileyonlinelibrary.com]

**TABLE 1** Collected raw data.

| Sensor | Data |
| --- | --- |
| Intel Realsense D435i RGB-D camera (front) | RGB-D Images, Stereo IR Images, IMU Readings |
| Intel Realsense D435i RGB-D camera (side) | RGB Images |
| Intel Realsense T265 tracking camera | Stereo Fisheye Images, Visual Odometry, IMU Readings |
| EMLID Reach RS-Plus RTK-GPS | GPS Coordinate (±4cm) |

Abbreviations: GPS, Global Positioning System; IMU, inertial measurement unit; IR, infrared; RTK, Real-Time Kinematic.

TABLE 2  Data categories.

| ID | Data category | Description |
| --- | --- | --- |
| a | Horizontal shadow | Shadow falls perpendicular to the direction of the crop row |
| b | Front shadow | Shadow of the robot falling on the image captured by the camera |
| c | Small crops | Crop rows at early growth stages (up to four unfolded leaves) |
| d | Large crops | Presence of one or many largely grown crops (more than four unfolded leaves) within the crop row |
| e | Sparse weed | Sparsely grown weed scattered between the crop rows |
| f | Dense weed | Weed grown densely among the crop rows where the interrow space is completely covered |
| g | Sunny | Crop row data captured in sunny weather |
| h | Cloudy | Crop row data captured in cloudy weather |
| i | Discontinuities | Missing plants in the crop row which leads to discontinuities in the crop row |
| j | Slope/curve | Images captured while the crop row is not in flat farmland or where crop rows are not straight lines |
| k | Tire tracks | Tire tracks from tramlines running through the field |

TABLE 3  Data classes (the numbers given in this table correspond to the indexes of combined field variations from different data categories).

| Data category | a | b | c | d | e | f | g | h | i | j | k |
| --- | --- | --- | --- | --- | --- | --- | --- | --- | --- | --- | --- |
| Horizontal shadow (a) | 1 | | | | | | | | | | |
| Front shadow (b) | | 7 | | | | | | | | | |
| Small crops (c) | 2 | 8 | 11 | | | | | | | | |
| Large crops (d) | 3 | 9 | 12 | 20 | | | | | | | |
| Sparse weed (e) | | | 13 | 21 | | | | | | | |
| Dense weed (f) | | | 14 | 22 | | | | | | | |
| Sunny (g) | | | 15 | 23 | 28 | 32 | | | | | |
| Cloudy (h) | | | 16 | 24 | 29 | 33 | | | | | |
| Discontinuities (i) | 4 | 10 | 17 | 25 | 30 | 34 | 36 | 39 | | | |
| Slope/curve (j) | 5 | | 18 | 26 | 31 | 35 | 37 | 40 | 42 | | |
| Tire tracks (k) | 6 | | 19 | 27 | | | 38 | 41 | 43 | | |

## 3.3 | Data preprocessing and annotation

The images were captured at 1280 × 720 resolution. These images were forced to resize to 512 × 512 resolution to allow seamless tiling of the output segmentation map (Ronneberger et al., 2015). The depth frames from the D435i camera have to be aligned to the RGB camera's viewpoint before combining a four-dimensional RGB-D image. An example of an RGB image of a sugar beet crop row and corresponding aligned depth frame is shown in Figure 4. The RGB images from the data set were then combined with corresponding depth images in grayscale color space. The RGB images and depth images are stored as two separate image files in the data set. The labels of crop rows are stored in an image file of an aligned crop row mask with a 6-pixel wide white line on black background. It is advisable to convert these label images to binary images before training a neural network. The coordinate data of the crop rows for each image is also stored as MATLAB data files along with the code to generate crop row masks of desired row width.

## 4 | METHODOLOGY

Our method comprises two main components: a crop row detection pipeline which predicts the central crop row which the robot will follow and a visual servoing controller which generates the velocity commands for the robot to follow. We propose a semantic segmentation approach for crop row detection based on U-Net

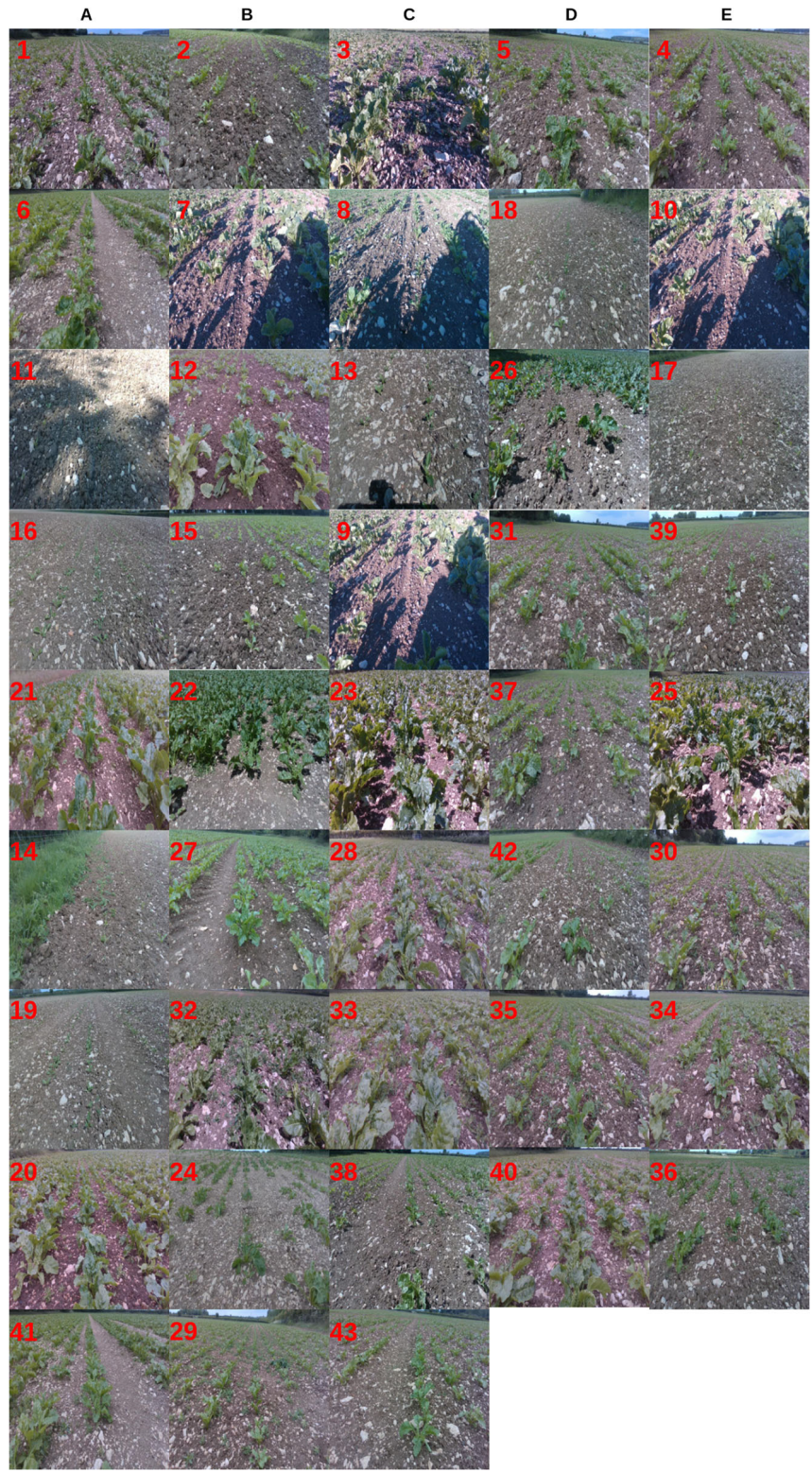

**FIGURE 2** Samples from data classes: (1–6) horizontal shadow, (7–10) front shadow, (2, 8, 11–19) small crops, (3, 9, 12, 20–27) large crops, (13, 21, 28–31) sparse weed, (14, 22, 32–35) dense weed, (15, 23, 28, 32, 36–38) sunny, (16, 24, 29, 33, 39–41) cloudy, (column E) discontinuities, (column D) slope/curve, (6, 19, 27, 38, 41, 43) tire tracks. [Color figure can be viewed at wileyonlinelibrary.com]

(Ronneberger et al., 2015). The labels were created in such a way that a crop row is represented as a single object rather than individual plants as shown in Figure 5. This representation will help the U-Net to predict the complete crop row despite the presence of discontinuities. The segmentation mask from U-Net is then used to identify the central crop row which the robot will follow. The visual servoing controller uses the angle and starting point of the detected crop row to drive the robot to the desired position. The overall architecture of our proposed visual servoing controller is illustrated in Figure 6.

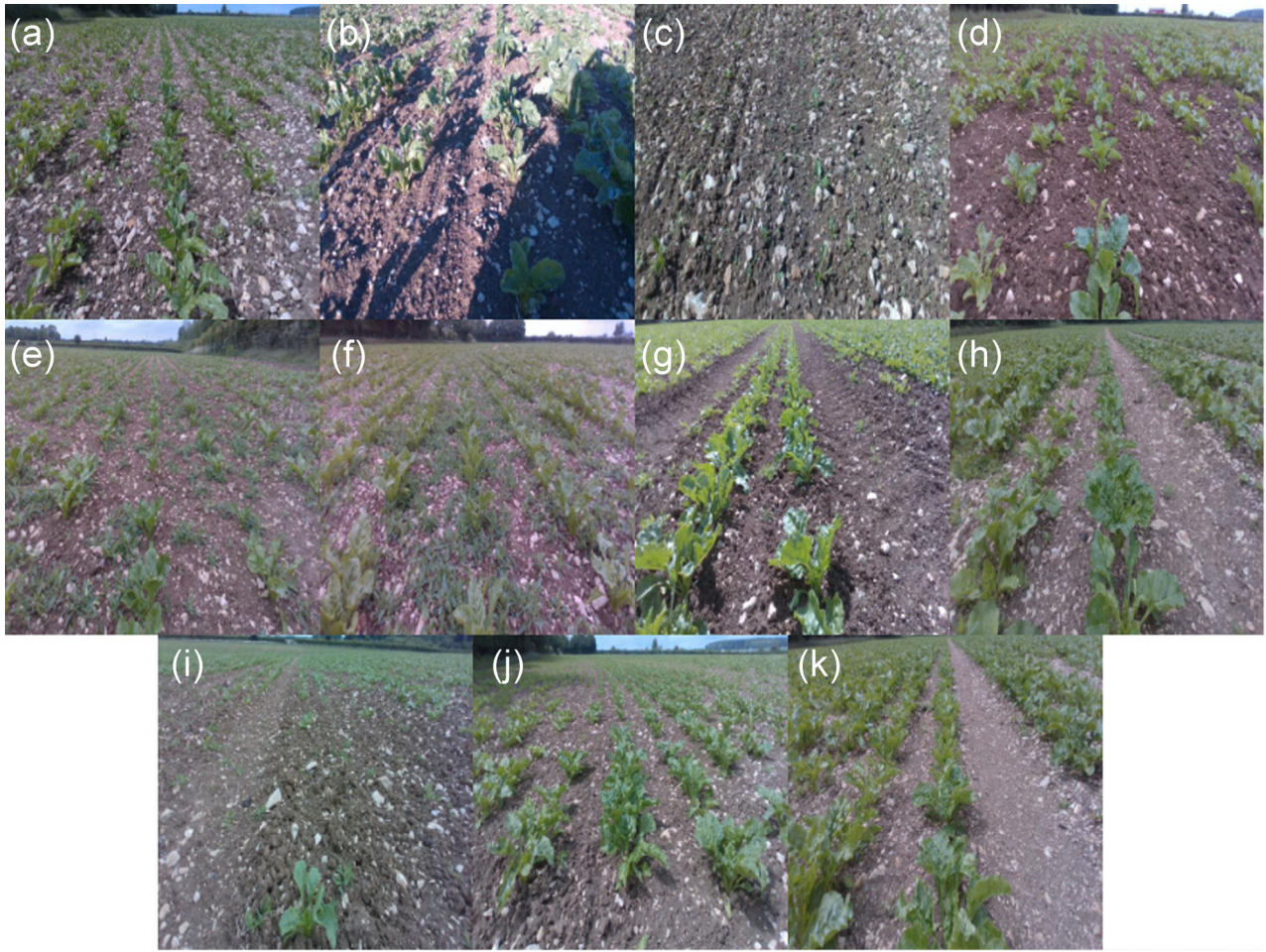

**FIGURE 3** Samples from 11 data categories. Each sample will exhibit a dominant field variation corresponding to the data category. (a) Horizontal shadow, (b) front shadow, (c) small crops, (d) large crops, (e) sparse weed, (f) dense weed, (g) sunny, (h) cloudy, (i) discontinuities, (j) slope/curve, and (k) tire tracks. [Color figure can be viewed at wileyonlinelibrary.com]

## 4.1 | Training U-Net

The U-Net model was trained with the Binary Cross Entropy (BCE) loss function with the Adam optimizer. The model was first trained only using RGB images and then trained again with RGB-D images to observe the contribution of depth data toward crop row detection. However, the addition of depth information did not account for a significant improvement in the predictions. Therefore, the model trained with RGB images was used.

Figure 7 shows the progress of U-Net learning to predict the crop rows with BCE loss function with RGB-only data. The model learns to predict the crops at five epochs of training. However, the predictions are only able to detect the regions where a crop lies in the image. Hence the crops which are not spaced closer are not recognized as complete crop lines. At 10 epochs the model becomes better at detecting the crop lines despite not being able to detect the line when there is a larger gap in the crop row. The model was still unable to detect the discontinuities in crop rows at 20 epochs. But the line predictions became more narrow and sharp. At around 40 epochs the model was able to detect and fill the gaps in crop rows predicting complete crop rows.

## 4.2 | Triangle scan method (TSM)

The TSM is a post-processing step for U-Net prediction. This method is used to determine the line parameters of the central crop row which the robot has to follow. The Realsense camera mounted on the Husky robot captures images of the parallel crop rows where the parallel crop rows appear to be converging to a point near the horizon of the field. This perspective distortion of the parallel crop rows due to camera placement could be exploited to detect the central crop row ($L$) accurately. The two endpoints of $L$ are assumed to be lying on the uppermost and the lowermost edges of the image, respectively. A triangular region of interest ($\Delta ROI$) is defined with three points: Anchor point ($A$), Begin point ($B$), and Cease point ($C$) as depicted in Figure 8. The $\Delta ROI$ is defined in such a way that it encompasses the region of the image where the central crop row usually resides. The points $B$ and $C$ are set to 190 and 350 (for a 512 × 512 image) after observing the occurrence of the lowermost point $L_{x2}$ of $L$ throughout the data set. The $\Delta ROI$ would ideally contain the pixels belonging to the central crop row, but it may contain pixel regions belonging to the neighboring crop rows as well. The algorithm has two steps to

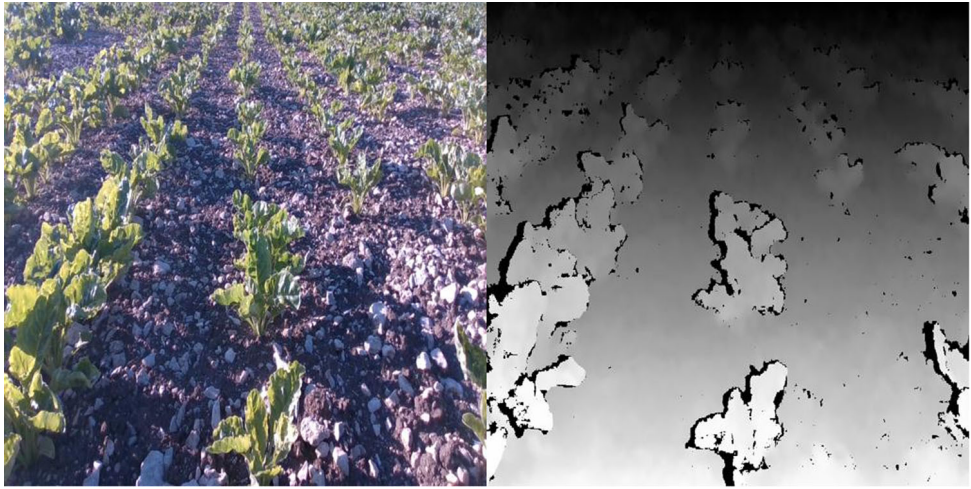

**FIGURE 4** RGB image example of a sugar beet crop row (left) and aligned depth image (right). [Color figure can be viewed at wileyonlinelibrary.com]

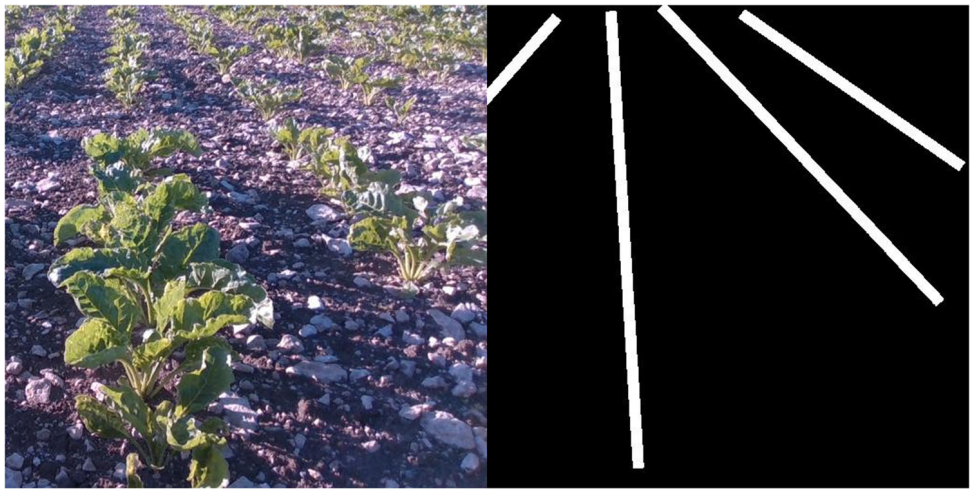

**FIGURE 5** Sample image (left) and respective ground truth label mask (right). Each crop row is marked with a white line with uniform width of 6 pixels in the ground truth mask. [Color figure can be viewed at wileyonlinelibrary.com]

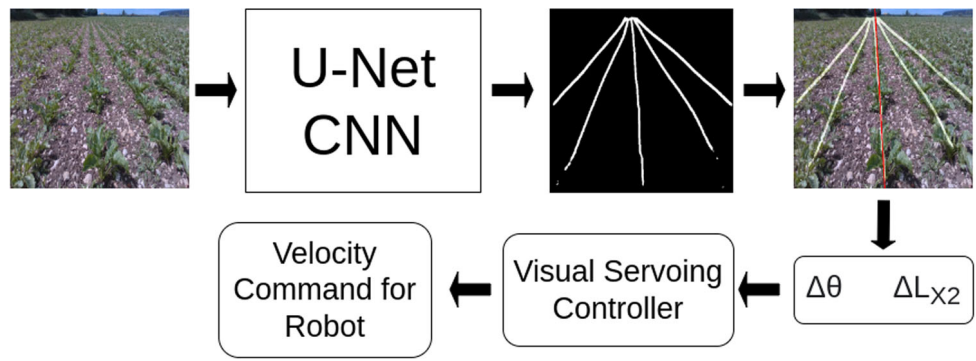

**FIGURE 6** The proposed crop row navigation architecture with U-Net CNN for crop row mask detection. The crop mask generated by U-Net CNN is used by a triangle scan method to predict a central crop row (Δθ, crop row angle error corresponding to the vertical axis; $\Delta L_{x2}$, positional error of the central crop row relative to image midpoint). CNN, convolutional neural network. [Color figure can be viewed at wileyonlinelibrary.com]

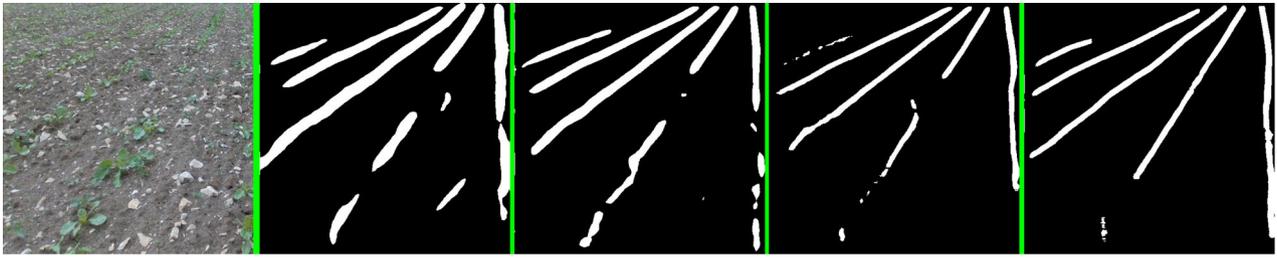

**FIGURE 7** Improvement of crop row mask prediction at 5, 10, 20, and 40 epochs, respectively (left to right). [Color figure can be viewed at wileyonlinelibrary.com]

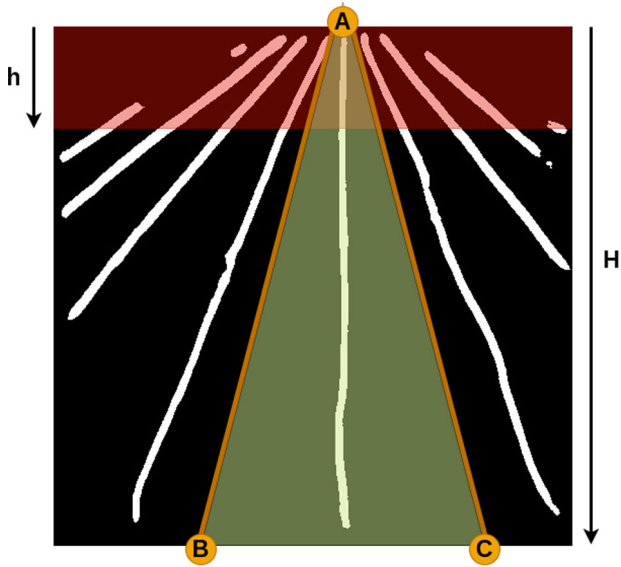

**FIGURE 8** Regions of interest for anchor scan and line scan. Anchor scan ROI, red; line scan ROI, green; H, height of the image; h, anchor scans ROI height. ROI, region of interest. [Color figure can be viewed at wileyonlinelibrary.com]

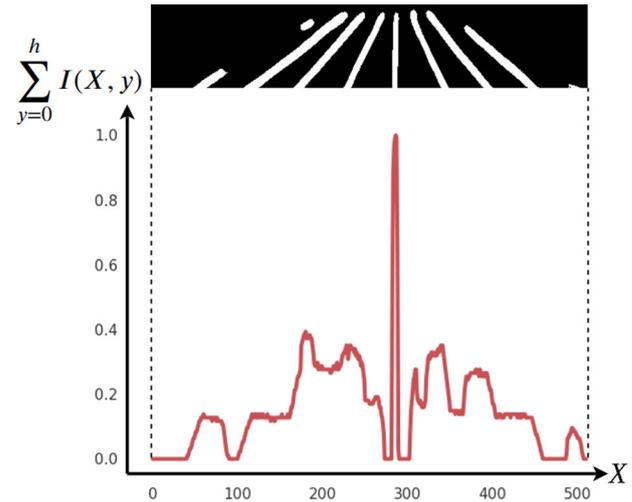

**FIGURE 9** Normalized sum curve for anchor scans (I, binary mask prediction from U-Net; X, vertical column positions in the rectangular ROI; y, pixel positions within a given vertical column). ROI, region of interest. [Color figure can be viewed at wileyonlinelibrary.com]

### 4.2.1 | Step 1—Anchor scans

The Anchor point (A) is determined for each image during this step. A horizontal rectangular strip from the top of the image with height $h$ is chosen as the ROI as shown in Figure 8. Equation: $h = sH$ denotes the relationship between $h$ and $H$ (height of the image) where $s$ is a scale factor between 0 and 1.

The numerical sum of each vertical pixel column in the selected rectangular ROI is then calculated and normalized. The peak point of the resulting sum curve is identified as the anchor point for the image. An example sum curve corresponding to the image in Figure 8 is given in Figure 9. The selection criteria to determine the upper point $L_{x1}$ of the central crop row is summarized in Equation (1) where $I$ is the binary mask prediction from U-Net, $X$ represents all the vertical column positions in the rectangular ROI and $y$ represents the pixel positions within a given vertical column.

determine the line. The first step is detecting the anchor point which is the upper endpoint of the crop row. The lower end of the crop row is detected in the second step. These steps are explained in Sections 4.2.1 and 4.2.2. The TSM is summarized in Algorithm 1.

| Algorithm 1. Triangle scan algorithm. |
|---|
| **Input**: Crop row mask prediction (I) from U-Net |
| **Output**: Line parameters of detected central crop row ($\theta$ and $L_{x2}$) |
| Detect anchor point: $L_{x1} \leftarrow \text{Arg Max}\left(\sum_{y=0}^{h} I(X, y)\right)$ |
| **if** maximum anchor scan sum ≤ threshold **then** |
|     set anchor point to preset value. |
| **end** |
| Detect lower point: $L_{x2} \leftarrow \text{Arg Max}\left(\sum_{y=0}^{H} I(X_{BC}, y)\right)$ |
| Calculate heading: $\Delta\theta \leftarrow \arctan\left(\frac{L_{x2} - L_{x1}}{\text{Image Height}}\right)$ |

$$L_{x1} = \text{Arg Max}\left(\sum_{y=0}^{h} I(X, y)\right). \quad (1)$$

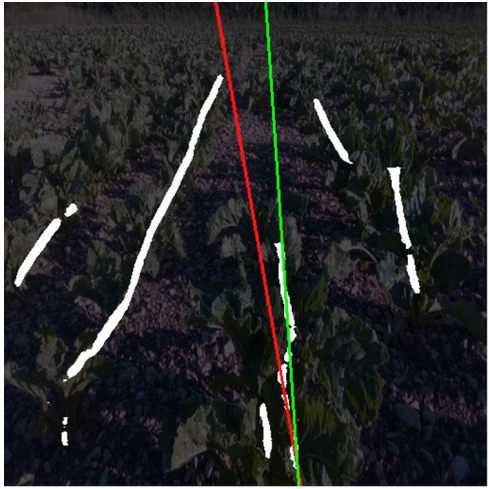

**FIGURE 10** An example of a poorly predicted crop row mask. (Red line) Central crop row prediction from anchor scans and (green line) central crop row prediction from predefined anchor point. [Color figure can be viewed at wileyonlinelibrary.com]

However, the anchor point is reset to a predetermined point (277 when the image width is 512) if the sum at the peak point is below a threshold value. The threshold value was set in such a way that the numerical sum of the pixel column must represent a vertical crop row with a minimum height of $0.4 \times h$. The predetermined anchor point is experimentally calculated by averaging the anchor points from a set of images equally drawn from all the data categories in the data set. This is done to avoid the algorithm from detecting false anchor points for poorly predicted crop row masks from U-Net. An example of such a poorly predicted crop row mask is given in Figure 10.

### 4.2.2 | Step 2—Line scans

The upper point of the crop line (anchor point) is determined during step 1. This step will determine the lower point of the central crop row. Let $P$ be any point on the line $BC$ of $\Delta ROI$ given in Figure 8. The sum of pixels along the $AP$ line is considered the scanner parameter for detecting the lower point of the central crop row. The selection criteria of the lower point $L_{x2}$ is expressed by Equation (2) where $X_{BC}$ represents all the points on line BC. An example sum curve corresponding to the image in Figure 8 is given in Figure 11.

$$L_{x2} = \text{Arg Max}\left(\sum_{y=0}^{H} I(X_{BC}, y)\right). \quad (2)$$

### 4.3 | Visual servoing experiment setup

"IBVS" is also commonly referred visual servoing controller in the literature as a classic controller for visual servoing. The IBVS

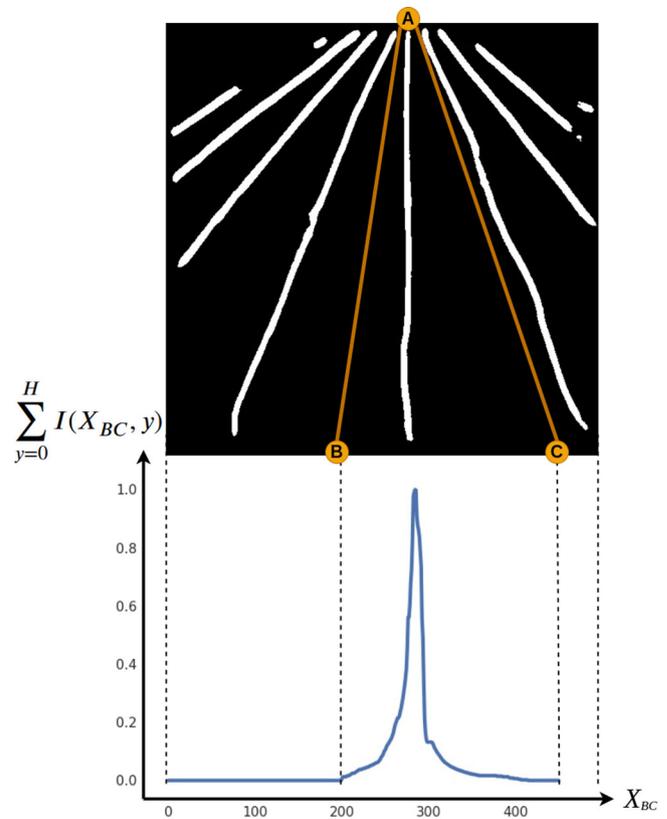

**FIGURE 11** Normalized sum curve for line scans ($l$, binary mask prediction from U-Net; $X_{BC}$, points on line BC; $y$, vertical position of a pixel within a given AP line). [Color figure can be viewed at wileyonlinelibrary.com]

implementation provided by Cherubini et al. (2008) is used in the baseline paper (Ahmadi et al., 2022). The IBVS controller equation is derived based on the state dynamics relationship defined in Equation (3).

$$\dot{s} = Ju = J_v v^* + J_\omega \omega. \quad (3)$$

The linear and angular velocity components of the robot are expressed by $u = [v^* \ \omega]^T$ and $J = [J_v \ J_\omega]$ is the image Jacobian matrix. The Jacobian matrix is calculated based on the camera intrinsics, the distance between the object and the camera, and the location of the desired feature point in the image (Chaumette et al., 2016). Equation (4) denotes the control equation for IBVS controller.

$$\omega = -J_\omega^+ (\lambda e + J_v v^*). \quad (4)$$

The error $e$ is $[\Delta L_{x2}, \theta]$ and $\lambda$ is a nonzero parameter that must be defined for a given control scenario. $-J_\omega^+$ is the Moore–Penrose pseudoinverse of $J_\omega$. Both IBVS and proportional controllers are feedback control systems with tuned parameters. However, IBVS controllers treat $V^*$ as a variable, allowing its value to be changed without modifying the $\lambda$ parameters of the controller. On the other hand, the gain of a proportional controller needs to be retuned when different values of V* are considered. However, the Jacobian matrix in the IBVS controller is calculated under the assumption that the mounting height and the forward-looking angle of the camera are fixed. This assumption is valid for most indoor scenarios and outdoor scenarios with minimal external disturbances. The fields in Lincoln, UK where our data set was collected presents an additional challenge due to the presence of significantly large rocks present in the interrow spaces as seen in data samples of Figure 3. This causes the robot and the camera to significantly disturb its pose while navigating through a crop row. Such disturbances emanate the Jacobian to be compensated to yield accurate control outputs in the IBVS controller. To this end, a proportional integral derivative (PID) controller is more suited to control the robot in such challenging environment due to its ability to compensate derivative and integral terms of the error while navigating the crop row. A proportional controller was chosen as the control algorithm for the robot in this controlled simulated environment with the intention of expanding it to a PID controller during real-world testing in the future.

Visual servoing is a method of using computer vision data to control a robot (Chaumette et al., 2016). The crop row detection method described in Section 4.2 could be used to implement a visual servoing-based proportional controller for crop row navigation. The objective of this experiment is to evaluate the impact of the proposed crop row detection algorithm on a visual servoing controller in a simulation environment free of uneven terrain. We have set up a simulated sugar beet field as shown in Figure 12. A real robot will alter its course due to external disturbances caused by uneven terrain. The simulated sugar beet field could be controlled to implement an even terrain which could be used to measure the impact of crop row detection toward visual servoing. A summary of the simulation parameters is presented in Table 4.

A proportional controller was implemented to perform visual servoing in the simulation environment. Figure 13 illustrates the positioning of the robot within crop row environment and the position of camera. The heading error, top and bottom points of the visible portion of central crop row in the world frame $R$ is denoted by $\theta_R$, $R_{x1}$, and $R_{x2}$ in Figure 13. The relationship between line parameters $L_R = [R_{x1}, R_{x2}]^T$ in world frame and line parameters in the image frame $L_I = [L_{x1}, L_{x2}]^T$ is expressed as $L_I = I_C {}^I T_R L_R$, where $I_C$ is the interaction matrix and ${}^I T_R$ is the homogeneous transform from world frame to camera frame. The controller will control the process variable $\omega$ which is the angular velocity of the robot around the z-axis. The forward velocity of the robot along the x-axis ($V^*$) is assumed to be a constant. The output from the TSM ($L_{x2}, \theta$) will act as a weighted input parameter to the proportional controller. The angular velocity $\omega$ is determined by the linear equation described in Equation (5) where $\alpha$ is the proportional gain and [$w_1$, $w_2$] are the contributing weights for angle and displacement errors of detected crop rows.

**TABLE 4** Simulation summary.

| Parameter | Value |
| --- | --- |
| Row spacing | 60 cm ± 0 |
| Seed spacing | 16 cm ± 0 |
| Plant height | 6 cm ± 3 cm |
| Plant orientation | 145° Randomized |
| Row length | 6 m ± 0 |
| Number of rows | 20 |

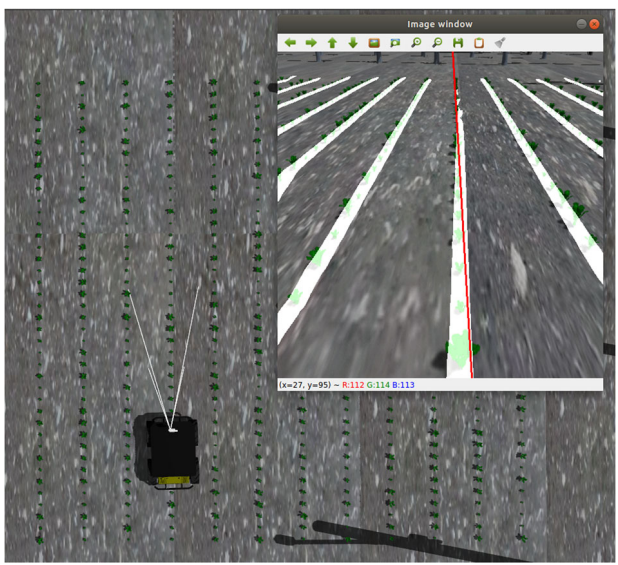

**FIGURE 12** Simulated sugar beet field for visual servoing (top right: robot camera view overlaid with U-Net prediction and TSM output). TSM, triangle scan method. [Color figure can be viewed at wileyonlinelibrary.com]

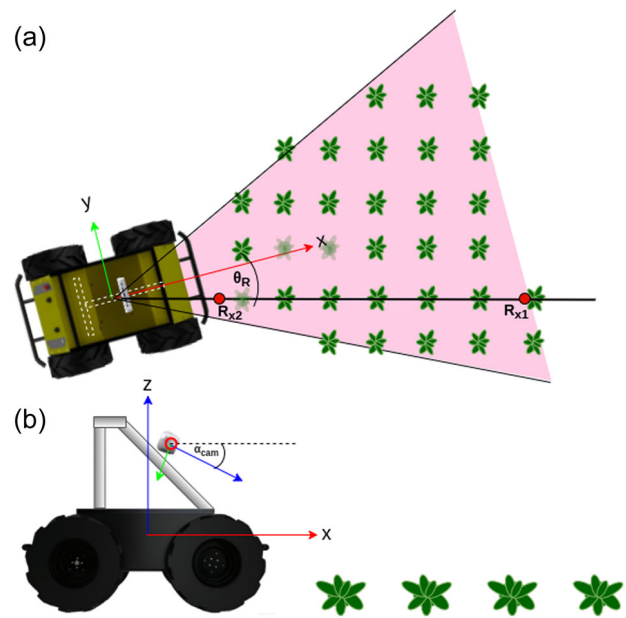

**FIGURE 13** Placement of Husky within a crop row (red, x-axis; green, y-axis; blue, z-axis). (a) Top view of Husky robot in a crop row and (b) side view of Husky robot in a crop row. [Color figure can be viewed at wileyonlinelibrary.com]

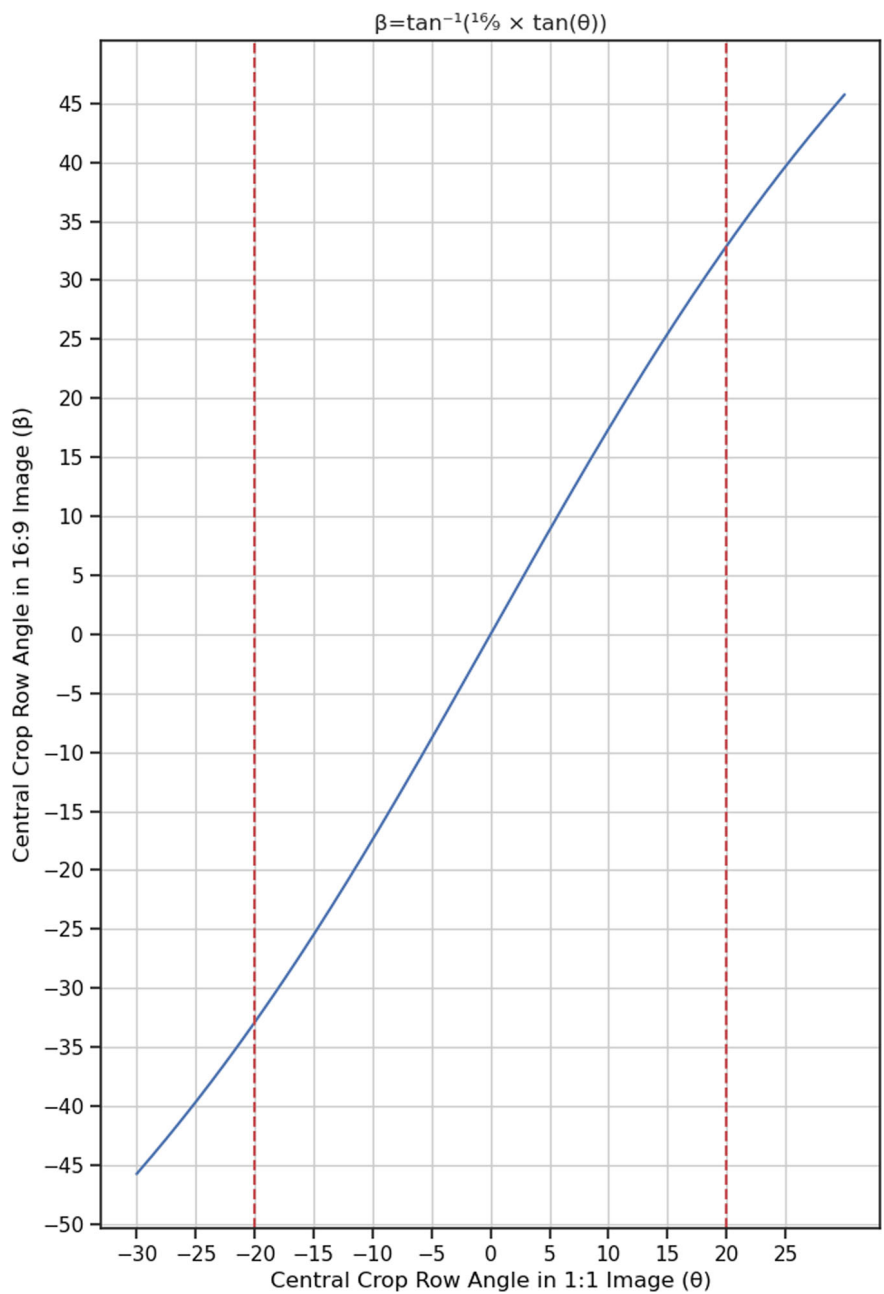

FIGURE 14 Distortion effect on central crop row angle due to resizing from 16:9 to 1:1 aspect ratio. [Color figure can be viewed at wileyonlinelibrary.com]

$$\omega = \alpha(w_1 \Delta\theta + w_2 \Delta L_{x2}). \quad (5)$$

The input RGB images used for the crop row detection algorithm were resized to a 1:1 aspect ratio from a 16:9 aspect ratio. The images were resized to meet the requirements of U-Net without compromising any information present in the original image (Ronneberger et al., 2015). The distortion introduced to the image through this resizing yields a different crop row angle in the detected central crop row relative to the original 16:9 ratio image. Figure 14 illustrates the effect of this resizing on the central crop row angle. We approximate this angular difference to a linear relationship within the limits of −20 to +20 of the $\theta$ angle. Therefore, that difference in predicted angle will not affect the overall visual servoing process since the $\alpha$ parameter in Equation (5) will compensate for any such linear distortions while ensuring the robot follows the real crop row while minimizing angular offsets. Any effect on the linear approximation of the nonlinear relationship will be reflected as a delay in the controller response.

## 4.4 | Crop row detection performance metric

The central crop row is detected using the TSM on the detected crop row mask from U-Net. Each detected crop row is parameterized by the angle it makes with the vertical direction $\theta$ and the position of the lowermost point of the line on the image $L_{x2}$. The absolute values are considered when calculating angle and displacement errors in Line $\Delta\theta$ and $\Delta L_{X2}$.

The overall performance of the crop row detection (ϵ) is quantified by a performance metric which equally weighs the line detection performance in terms of $\Delta\theta$ and $L_{x2}$ as expressed in Equation (6). $N$ is the number of images being tested. $\Delta\theta_{max}$ and $\Delta L_{x2,max}$ are the maximum detected errors for each $\Delta\theta$ and $L_{x2}$, respectively. All the angles are measured in degrees and displacements are measured in pixels.

$$\epsilon = 1 - \sum_{i=0}^{N} \frac{1}{2N}\left(\frac{\Delta\theta_i}{\Delta\theta_{max}} + \frac{\Delta L_{x2,i}}{\Delta L_{x2,max}}\right). \quad (6)$$

## 5 | RESULTS AND DISCUSSION

The proposed crop row detection pipeline was tested with a test data set of 430 images which has 10 images per data class listed in Table 3. Images in any given class are nonconsecutive frames which are drawn from the same crop row or multiple crop rows. We have developed a Gazebo simulation with a realistic sugar beet field to test the performance of our method in driving a robot along a crop row. The U-Net model used in this simulated experiment is trained on a data set with images from a simulation. The process of obtaining the simulated data set is explained in our prior work (de Silva et al., 2022).

### 5.1 | Baseline

Our approach to navigating an arable field with crop rows consists of two parts: crop row detection algorithm and visual servoing controller. The work done by Ahmadi et al. (2022) shares the same goal with similar objectives. Their approach to crop row detection is based on color-based crop segmentation followed by a crop centers detection algorithm. They use the detected crop centers to identify the central crop row while predicting crop rows with the least square fitting. However, their implementation requires certain limitations in camera positioning that avoids the visibility of the vanishing point in the captured images. The ROI parameters are set to meet these requirements in the CRDLD data set during the experiments. Their method was chosen as the baseline since they are using classical computer vision techniques rather than deep learning-based segmentation. In contrast, our deep learning-based segmentation approach relies on uniform crop row labels which are independent of crop type, growth stage, and discontinuities. Therefore, this comparison will highlight the intrinsic limitations of classical color-based segmentation methods and emphasize the advantages of deep learning-based approaches in generalizing the crop row detection. Their work is sufficiently recent that they have considered the pre-existing crop row detection approaches to develop their algorithm. This enables us to compare our algorithm with a sufficiently novel and recent baseline which supersedes the earlier work in vision-based crop row detection.

### 5.2 | Categorical evaluation

The U-Net model used in this experiment was trained on a data set of 1075 images which has 25 images per data class listed in Table 3. The peak validation IoU of RGB-only model reported 22.5% while the RGB-D model reported 31.75%. The qualitative difference in the predictions of these two models is insignificant despite the higher IoU in RGB-D model as shown in Figure 15. Addition of depth information to the crop row detection pipeline also increases the computational time for crop row detection. Therefore, the RGB-only model was chosen to predict crop row masks for the TSM. The reported IoU values would be considered as below-average performance for a typical semantic segmentation task in general computer vision applications. However, our postprocessing algorithm can recover accurate crop rows for navigation regardless of the lower IoU score. The individual IoU score in our test data set varied between 10% and 50%. However, the $\Delta\theta$ throughout the IoU range was below 2° and $\Delta L_{X2}$ was below 20 pixels in image space within the IoU range. The performance of our method is compared with the baseline: multicrop row detection method presented in Ahmadi et al. (2022). The result of categorical evaluation for our method and the baseline is summarized with three indicators: $\Delta\theta$, $\Delta L_{X2}$, and ϵ in Table 5. The table given in Appendix A presents the same evaluations calculated per each data class. The maximum errors: $\Delta\theta_{max}$ and $\Delta L_{x2,max}$ were recorded from the baseline method in categories 42 and 6, respectively.

The results indicate that our method is on average 34.62% better than the baseline in terms of detecting crop rows according to the ϵ metric. Our method can predict crop row angle with 2.86° less error, which is 61.5% improvement from the baseline. The displacement detection error of our method is 38.66 pixels lesser, which is 76.33% improvement from the baseline for an image with a width of 512 pixels. The baseline algorithm failed to detect any crop rows in 11.86% of the images in the test data set while our method could detect crop rows in all the images in the test data set. The detection results for baseline in Table 5 are calculated only considering the successful cases of detection.

The detection failures in the baseline method were recorded from 24 classes in Table 3. However, it was noted that 41.18% of the failures were recorded in three classes: 6, 27, and 41. All of these three classes were having tram lines present in the images. This indicates that the baseline method is dependent on equal spacing among crop rows to accurately predict crop rows. The average baseline ϵ for these three classes was 44.82% which is below the average ϵ for the baseline method. This average ϵ value suggests that the baseline method was performing poorly in these classes successful detection cases. Our method has been able to predict crop rows with average $\Delta\theta$, $\Delta L_{x2}$, and ϵ of 1.45°, 8.2 pixels, and 85.55%, respectively, for these three classes.

Curved crop rows are a challenging scenario for both algorithms since both algorithms approximate the crop row to a straight line. However, the apparent curvature of the line is only visible at the far end of the crop row due to the perspective distortion caused by camera placement. Therefore the near end of the crop row appears to be a straight line while the far end of the crop row appears to be a curved line in curved crop row images. We assume that a successful crop row detection algorithm should accurately predict the crop row at the near end since that prediction is important for the immediate control of the robot in a visual servoing controller. The ground truth lines of the curved crop rows were labeled as straight lines adhering to this assumption. The outputs of both our method and the baseline predict the central crop row aligning to the near-end straight line segment of the crop row as depicted in Figure 16. The average baseline $\epsilon$ for curved crop rows was 51.12% and the average $\epsilon$ for our method was 82.45%. Although our method performed significantly better than the baseline, its performance was below average compared to the overall performance of its own.

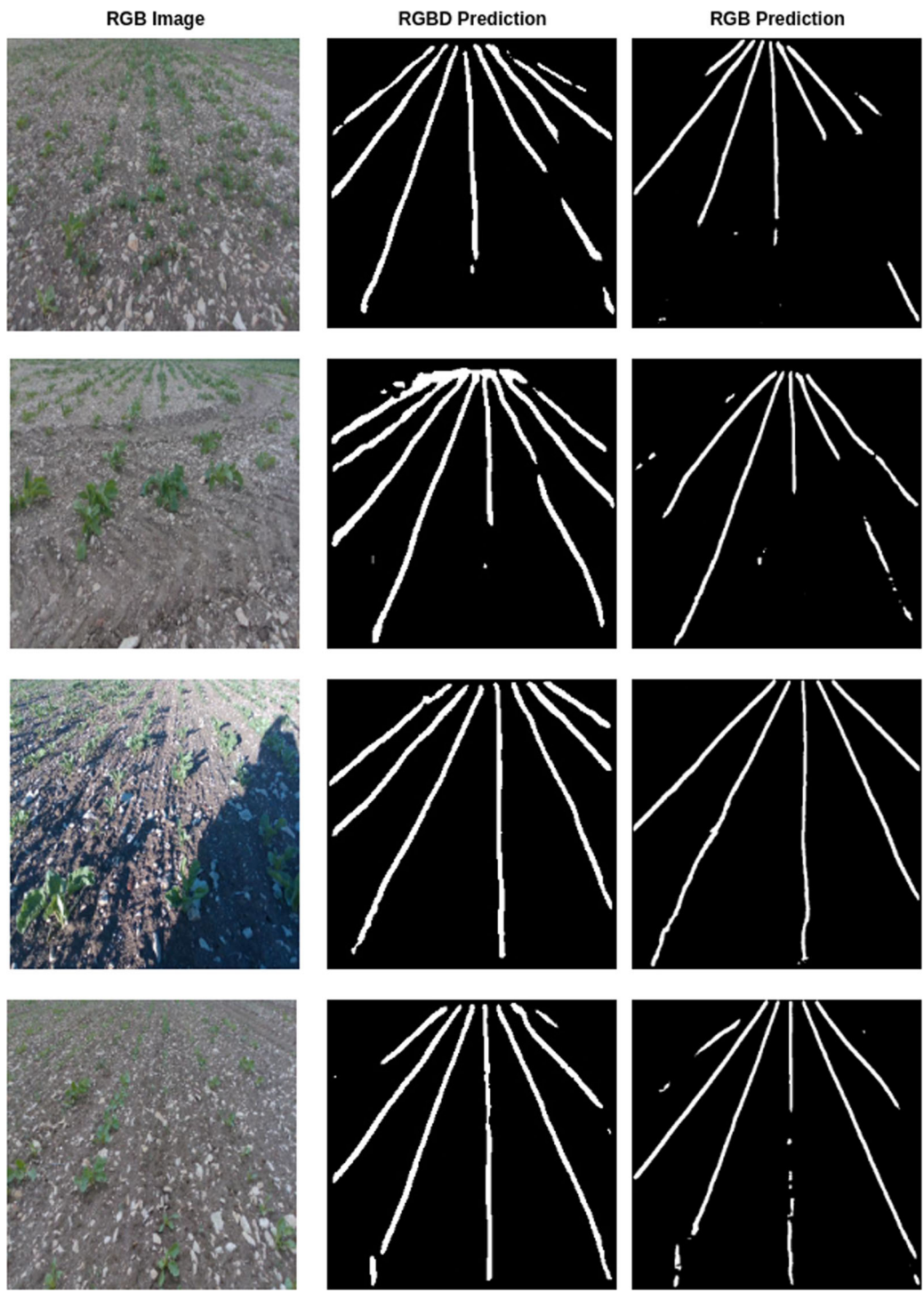

**FIGURE 15** Comparison of crop row masks generated by U-Net models trained with RGB-D images and RGB images. RGB input image (left), RGB-D U-Net prediction (middle), and RGB U-Net prediction (right). [Color figure can be viewed at wileyonlinelibrary.com]

Figure 17 shows a crop row detection instance of both algorithms at an edge of a field. It can be noticed that the baseline algorithm identifies the grass patch on the left as another plant and predicts the central crop row based on its detection. However, our method has been able to accurately identify the crop row while not detecting any false regions in the segmentation mask. This example highlights the capability of deep learning-based segmentation in crop row detection algorithms when compared to the traditional color-based segmentation approaches. Deep learning-based segmentation will predict the crop row masks while detecting the correct crop type to form the crop row. Deep learning-based segmentation is robust against different field conditions and anomalies. Therefore, deep learning-based segmentation is more suitable for real-world implementations of crop row detection.

The presence of weeds is categorized into two main groups depending on the density of their spread in the field. The images with few weeds spread across the interrow space are called "Sparse Weed" and images with the interrow space completely occupied by weeds are called "Dense Weed" as explained in Table 2. The average $\epsilon$ for image classes under "Dense Weed" category for the baseline and our method were 48.5% and 82.25%, respectively. The average $\epsilon$ value for both of these categories was below average indicating that the presence of high-density weeds in arable fields is a challenging field condition for crop row detection algorithms. Conversely, the "Sparse Weed" category is not a challenging field condition for crop row detection algorithm since both the baseline and our method performed above average in this category. The average $\epsilon$ for baseline and our method in this category were 56.9% and 86.88%, respectively.

The growth stage is a constantly changing field condition. The early growth stages have around four unfolded leaves which are small and the later growth stages have 6 or more unfolded leaves. The available free space between two rows is higher in the early growth stages due to smaller leaves. Some of the interrow space will be occupied by the leaves in later growth stages. Our method recorded an average $\epsilon$ of 75.33% and 86.44% for early and later growth stages, respectively. The baseline method recorded 54.03% average $\epsilon$ for early growth stages and 53.59% average $\epsilon$ for later growth stages. Our method performed better than the baseline in both the "Small Crops" and the "Large Crops" categories. The earlier growth stages are one of the most challenging field variations for our method. Our method could accurately predict crop rows in most of the test cases in "Small Crops" category despite the overall lower score. However, there were few images with incorrect anchor point detection leading to an overall lower score. The predicted crop row masks in these images had poorly predicted row structure at the far end of the crop

**TABLE 5** Crop row detection errors in each category ($\Delta\theta$, average angular error; $\Delta L_{X2}$, average; $L_{X2}$, error; $\epsilon$, overall crop row detection performance; $B$, baseline).

| Data category | $\Delta\theta_B$ | $\Delta\theta$ | $\Delta L_{X2,B}$ | $\Delta L_{X2}$ | $\epsilon_B$(%) | $\epsilon$(%) |
|---|---|---|---|---|---|---|
| Horizontal shadow | 4.96 | 1.98 | 58.20 | 14.79 | 46.92 | 82.13 |
| Front shadow | 4.08 | 1.02 | 39.38 | 9.27 | 59.67 | 90.13 |
| Small crops | 4.64 | 1.86 | 45.02 | 13.91 | 54.03 | 83.2 |
| Large crops | 4.45 | 1.55 | 49.13 | 10.46 | 53.59 | 86.44 |
| Sparse weed | 3.54 | 1.29 | 54.71 | 13.48 | 56.91 | 86.88 |
| Dense weed | 5.11 | 1.97 | 51.89 | 14.67 | 48.51 | 82.26 |
| Sunny | 4.27 | 1.4 | 50.26 | 9.48 | 54.24 | 87.73 |
| Cloudy | 4.36 | 1.35 | 54.45 | 10.03 | 52.00 | 87.87 |
| Discontinuities | 4.87 | 2.07 | 48.91 | 12.63 | 51.12 | 82.45 |
| Slope/curve | 4.03 | 1.65 | 41.59 | 10.43 | 59.13 | 85.88 |
| Tire tracks | 4.68 | 1.59 | 67.84 | 12.10 | 44.82 | 85.55 |
| **Average** | **4.51** | **1.65** | **50.65** | **11.99** | **50.65** | **85.27** |

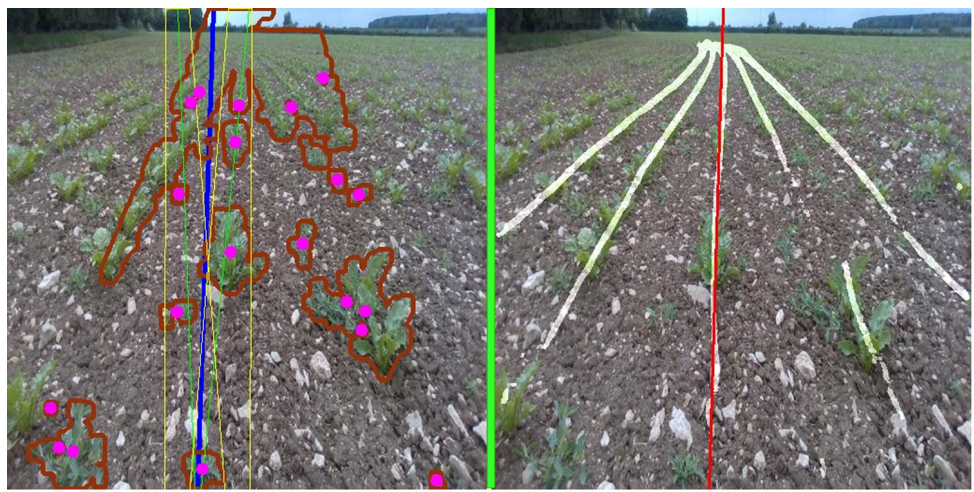

**FIGURE 16** Crop row detection in curved crop rows. (Left) Baseline (yellow, bounding box for crop row; green, predicted crop row; blue, navigation line; pink, predicted plant locations; brown, segmented plants) and (right) our method (white, crop row segmentation mask; red, navigation line). [Color figure can be viewed at wileyonlinelibrary.com]

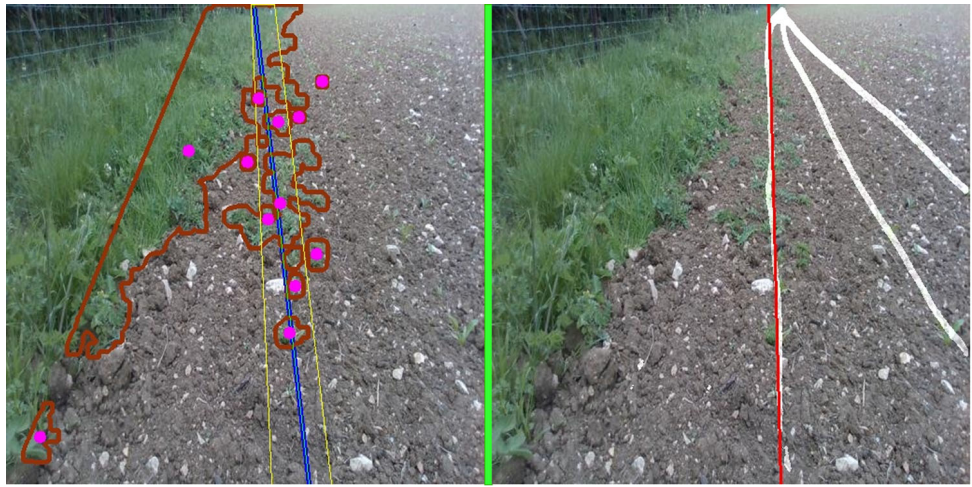

**FIGURE 17** Crop row detection at edges of the field. (Left) Baseline (yellow, bounding box for crop row; green, predicted crop row; blue, navigation line; pink, predicted plant locations; brown, segmented plants) and (right) our method (white, crop row segmentation mask; red, navigation line). [Color figure can be viewed at wileyonlinelibrary.com]

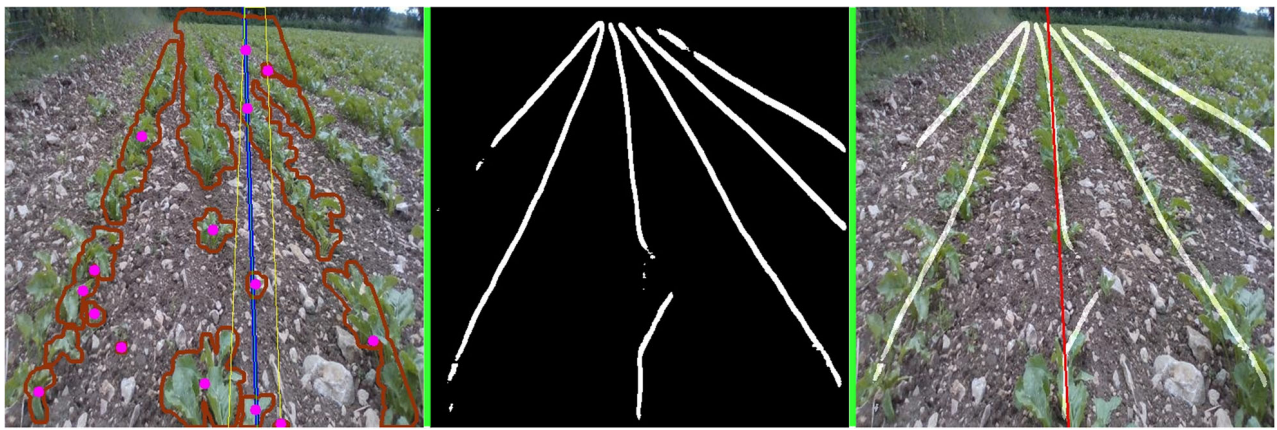

**FIGURE 18** Crop row detection with discontinuities. Baseline (yellow, bounding box for crop row; green, predicted crop row; blue, navigation line; pink, predicted plant locations; brown, segmented plants); middle, U-Net prediction; and right, our method (white, crop row segmentation mask; red, navigation line). [Color figure can be viewed at wileyonlinelibrary.com]

row, leading to incorrect anchor scan detection. The imbalance of the data set in terms of grow stage variation could be a reason for such inconsistent crop row mask predictions.

Discontinuities in crop rows are also another challenging field condition for both methods. Both the methods were performing at below average $\epsilon$ values since both methods rely on the robust structure of the crop rows. The average $\epsilon$ for baseline and our method for images with discontinuities were 44.89% and 75.15%, respectively. The discontinuities in the crop rows cause missing plant centers in the baseline method and thus lead to inaccurate results. The U-Net model used in our method will predict the crop row mask for the entire line when small discontinuities are present where one or two plants are missing in the line. However, it fails to detect and fill the line segments for large discontinuities. The TSM is able to predict the central crop row with reasonable accuracy even with such broken lines. Figure 18 shows a detection scenario with discontinuities.

Class 31 presents an interesting result where both the baseline and our method exhibit similar $\epsilon$ performance scores. While the baseline performs above average $\epsilon$, our method has the second least $\epsilon$ value in this class. Nine out of the 10 test images in this class performed well within the usual error margins. One of the images in the test set predicted a neighboring crop row as the central crop row as shown in Figure 19, leading to a lower class performance score. A similar situation has caused class 2 to become the least performing class in our crop row detection algorithm. A common feature in both of those failed images is that the $L_{x2}$ point of both of these images was below 200 which causes the central crop row to fall outside the scanning range (BC line) of the line scanning step in our crop row detection algorithm. The scanning range points $B$ and $C$ were set by observing the usual region where the central crop row resides in all the images. These two failure instances were borderline cases where the central crop row is not residing within usual limits. Upon further investigation into the images, it was revealed that the human controller have lost control of the robot for a few seconds during data collection and it causes the camera to capture these images with unusual orientation. The effect of these failures could be ignored with the assumption that an autonomous controller would not lose the control of the robot to such an extent although these failures reduced the overall performance scores of our algorithm in their respective classes. Given the assumption that the robot enters the crop row with a maximum angular error of 20°, our experiments in Section 5.3 show that our visual servoing controller will not drive the robot to a position where the $L_{x2}$ point of the central crop row is outside the BC line unless the robot's heading is altered by an external force. Therefore, such a scenario will not occur when the robot is being driven by the visual servoing controller. However, this failure gives an insight into the codependent nature of our crop row detection algorithm and visual servoing controller. Any large angular errors caused by the visual servoing controller will lead the camera to capture the images at unusual orientations which will lead to false central line predictions causing the robot to shift to the neighboring crop row. Such a scenario was intentionally triggered and the robot eventually adjusted and followed the neighboring crop row until the end. However, the robot damages some of the crops in the traversing crop row while shifting.

## 5.3 | Visual servoing experiment

The visual servoing setup described in Section 4.3 was used to devise an experiment to observe the impact of the crop row detection algorithm on visual servoing. The robot was placed at the starting point of each crop row such that the coordinates of the robot coincide with the coordinates of the starting position of the crop row. The heading direction of the robot was randomly assigned to be ±20° from the direction of the crop row. The crop row was modeled to be in a perfect straight line. The robot was allowed to follow the line until it reaches the end of the crop row (no crops are visible on the top half of the image). The angle and displacement errors of the robot were measured. Any motion that occurs on the robot is considered to be caused by the visual servoing controller since the field is modeled to mimic an even terrain. The initial heading of the robot was set to a random positive orientation for the first 10 trials and it was set to a random negative orientation during the last 10 trials. Each trial took approximately 140 frames to follow the 6-m crop row. Few examples of angle error $\Delta\theta$ and displacement error $\Delta L_{x2}$ variations during a selected trials are given in Figures 20 and 21. The red color curve indicates the average $\Delta\theta$ and $\Delta L_{x2}$ for all the trials.

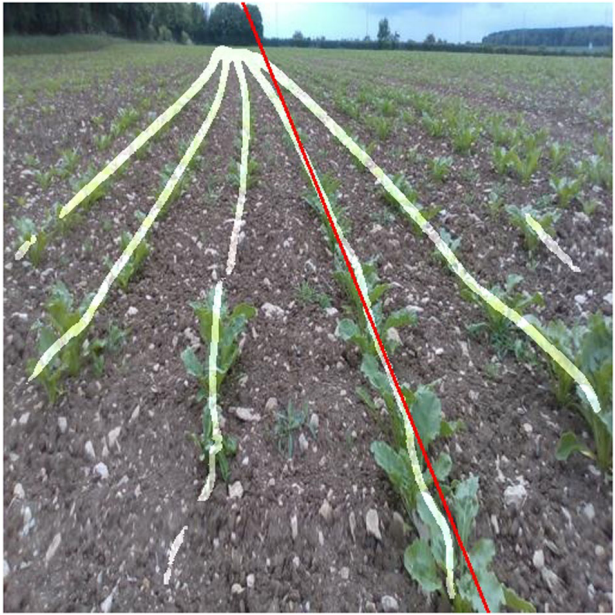

**FIGURE 19** False central crop row prediction in class 31 (slope/curve and sparse weed) (white, crop row segmentation mask; red, navigation line). [Color figure can be viewed at wileyonlinelibrary.com]

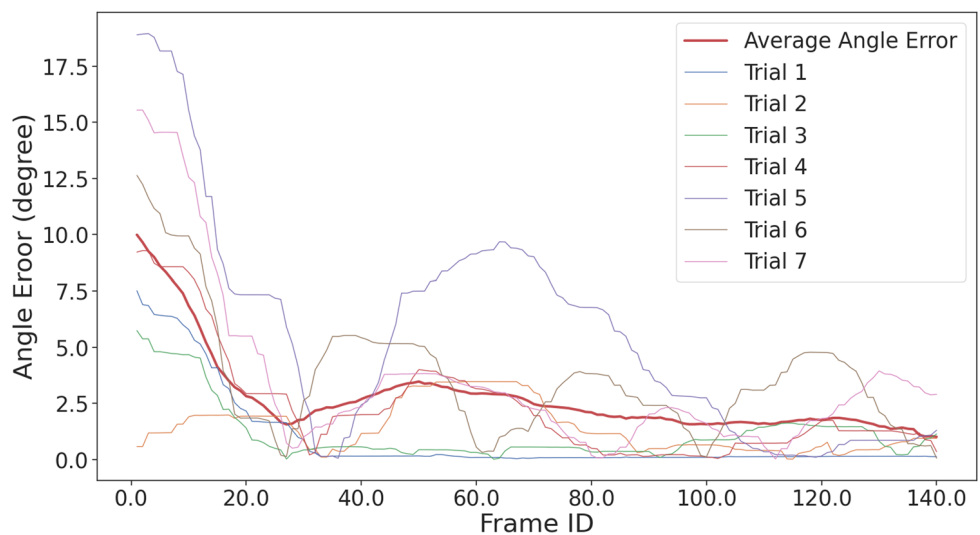

**FIGURE 20** Angle error $\Delta\theta$ variation during a selected trials of visual servoing (average curve indicate the average error on all 20 trials). [Color figure can be viewed at wileyonlinelibrary.com]

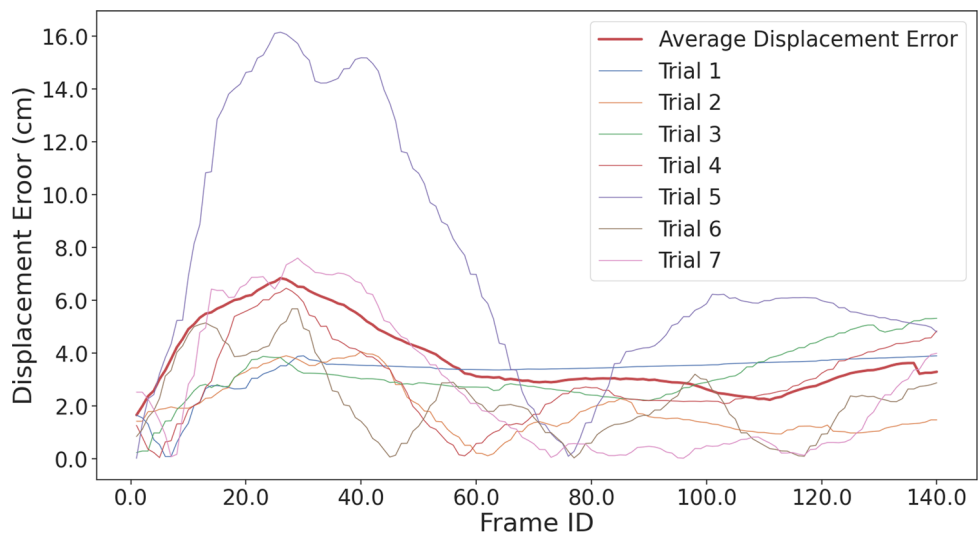

**FIGURE 21** Displacement error ΔL$_{x2}$ variation during selected trials of visual servoing (average curve indicate the average error on all 20 trials). [Color figure can be viewed at wileyonlinelibrary.com]

**TABLE 6** Hyperparameters related to triangle scan algorithm.

| Parameter | Description |
| --- | --- |
| Begin point (B) | Starting point of the bottom edge of ΔROI described in Section 4.2 |
| Cease point (C) | End point of the bottom edge of ΔROI described in Section 4.2 |
| Anchor scans ROI scale factor (s) | Scale factor s in Equation h = sH described in Section 4.2.1 |
| Anchor scan threshold | Threshold value for Equation (1) to validate anchor scans |

Abbreviation: ROI, region of interest.

The settling time is defined as the number of frames taken by the algorithm to first reach 2° error. The settling time appears to be 20 frames in the example given in Figure 20. The overall average settling time for all 20 trials was 23 frames. The Δθ stayed below 2° in this example. However, the Δθ swung back to above 2° errors in some trials. The displacement error increases in the beginning and then settles due to the controller overshoot while attempting to correct the angle error. This could be corrected using a PID controller. The average of absolute Δθ and ΔL$_{x2}$ were found to be 2.15° and 3.56 cm, respectively. Similar values were observed when the initial heading of the robot was set to 0° to align with the crop row This measure could be considered as the average error of the visual servoing controller caused by the crop row detection algorithm.

## 5.4 | Hyperparameter analysis

The robustness of the hyperparameters of the proposed crop row navigation algorithm illustrated in Figure 6, is examined in this section. The hyperparameters related to U-Net CNN and crop mask prediction is evaluated in our previous work de Silva et al. (2022). The hyperparameters of the triangle scan algorithm which we introduce in this work are listed in Table 6.

### 5.4.1 | Begin point (B) and cease point (C)

The points B and C enclose the scanning range for L$_{x2}$ in Equation (2). The ground truth values of the data set were examined to set ideal values for B and C. The frequency of the bottom point of the central crop row in the data set is plotted in Figure 22. The L$_{x2}$ point in data set is considered as an outlier if the frequency is less than 5. The minimum L$_{x2}$ point excluding such outliers is chosen as point B for the triangle scan algorithm. The maximum L$_{x2}$ point excluding the outliers is chosen as point C. Widening of the B–C range would increase the execution time of the triangle scan algorithm despite a broader B–C range could include a more comprehensive scanning range. 2.33% of test data set was found to have L$_{x2}$ points lying outside the scanned B–C range. The benefit of shortening the execution time was considered to outweigh the disadvantage of having such outliers. The B, C points could be set to the bottom left and bottom right corners of the image in the absence of such ground truth data to determine the B, C values.

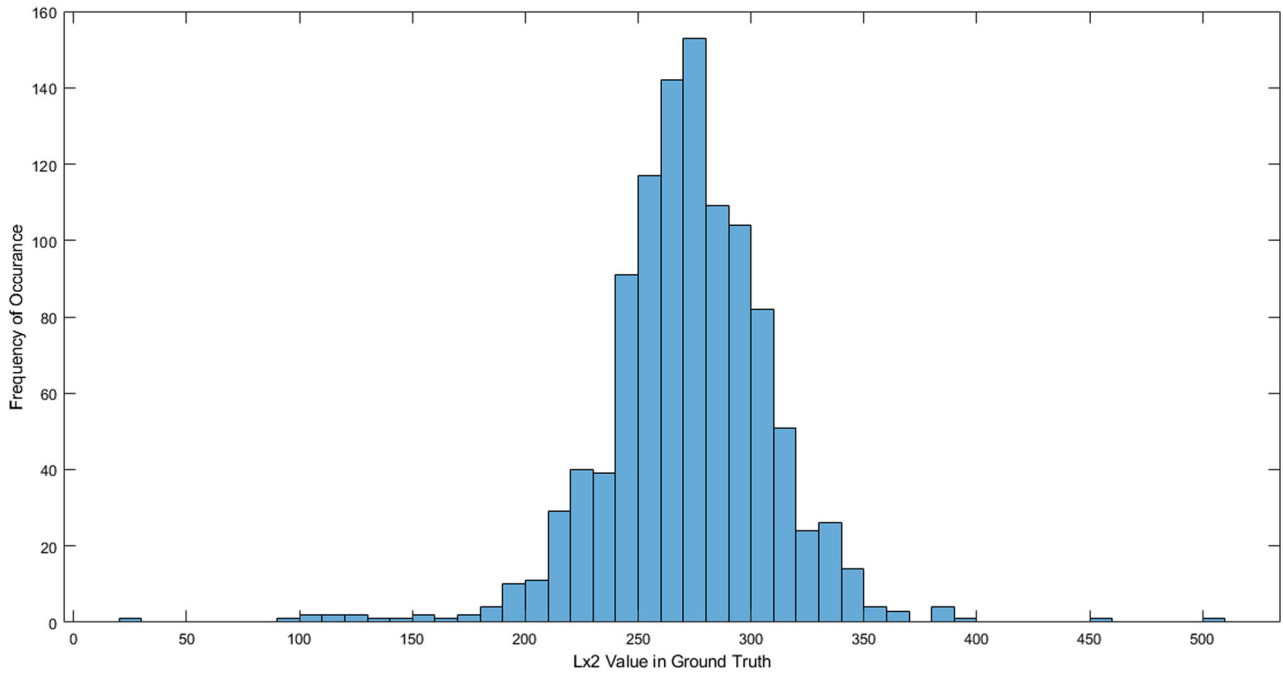

**FIGURE 22** Histogram plot for the occurrence of $L_{x2}$ in ground truth data. [Color figure can be viewed at wileyonlinelibrary.com]

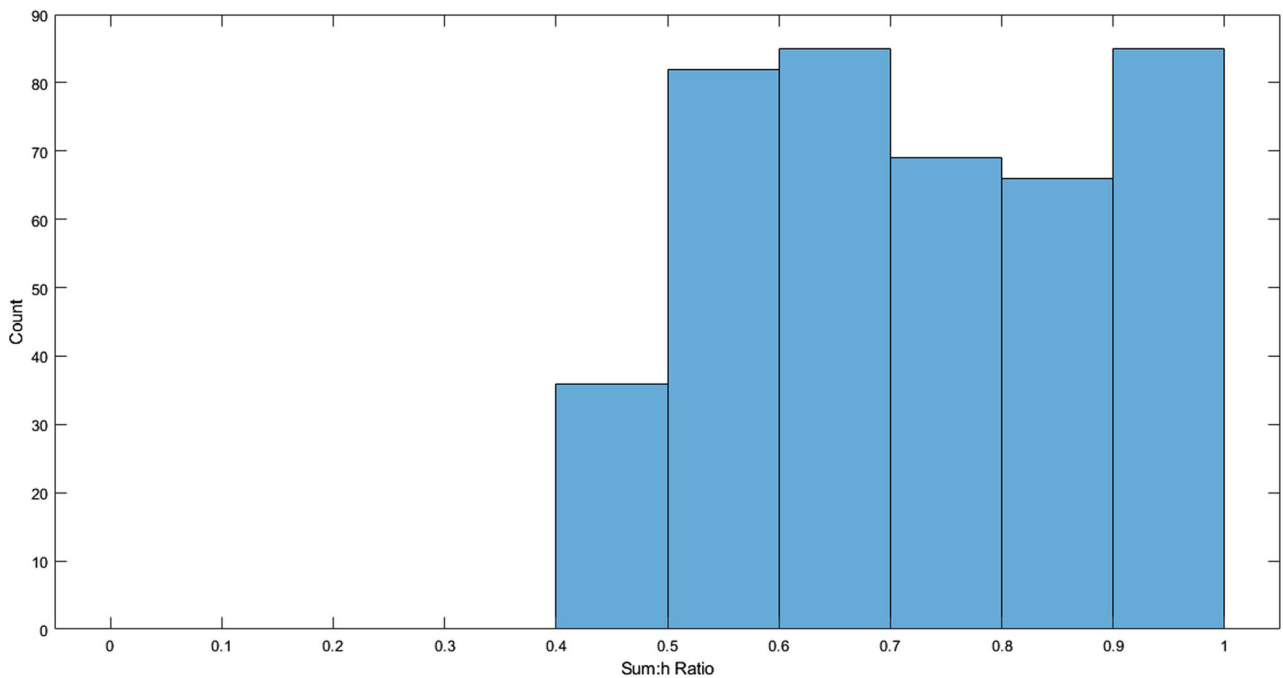

**FIGURE 23** Histogram plot for $\sum_{y=0}^{h} I(X, y)$ to $h$ ratio in predicted crop row masks. [Color figure can be viewed at wileyonlinelibrary.com]

### 5.4.2 | Anchor scans ROI scale factor and threshold

The anchor scans ROI scale factor is the $s$ value referred in $h = sH$ described in Section 4.2.1. This scale factor defines the height of the rectangular ROI in which the anchor points were scanned. This value was set to 0.2 based on qualitative analysis of the image data. The anchor point of the central crop row was at least within the top 20% of the image across all images in the data set. A smaller $s$ value could yield a shorter execution time for the anchor scans step of the algorithm. It was also noted that every 10% increment in the $s$ value increases the anchor scan execution by 1 ms.

Anchor scan threshold is the margin value for the ratio between scanner parameter in Equation (1) and $h$ to validate an anchor scan. The histogram given in Figure 23 was plotted to visualize the distribution of this ratio across the test data set. The minimum recorded ratio was identified as 0.4. Therefore any anchor scan which yields less than 0.4h in Equation (1) is replaced with a predefined anchor point.

## 6 | FUTURE WORK

The experimental results from crop row identified the challenging field conditions for crop row detection. A new data set will be composed of adding more images from the most challenging field conditions to improve crop row mask prediction. Addition of a dense growth stage classification would enable more closer observation of the effect of plant growth on the proposed vision-based navigation system. The points B and C were predefined to match the images in our data set during the testing of the algorithms. We will develop a method to identify and dynamically adjust these points while the robot is autonomously following the crop rows. The visual servoing controller was tested on a simulator to inspect its performance. We will conduct real-world experiments to validate the performance of the visual servoing controller during the next crop season. The Husky robot was able to follow a crop row up to 6 meters during early-stage real-world field trials. However, the controller must be further optimized to execute real-world trials more accurately. The work presented in this paper was only focused on following a single crop row in an arable field. We will extend our anchor scans algorithm to detect the EOR (End of crop Row) and subsequently develop a row-switching algorithm to fully navigate a crop row field.

## 7 | CONCLUSION

We have presented a novel and effective way of predicting the crop rows and following the central crop row in a sugar beet field. Our method only uses RGB images as input and it can predict crop rows under varying field conditions with reasonable accuracy than the baseline. Our method can detect the central crop row with an average angular error of 1.65° and an average displacement error of 11.99 pixels within the given image space. We managed to improve the angle error by 61.5% and displacement error by 76.33% than the baseline while maintaining a 100% success rate in detection. We have identified that earlier grow stages and having large discontinuities in the crop row are the most challenging field condition for our algorithm to detect crop rows in a real-world environment. The simulation experiment conducted on visual servoing concluded that our method can follow the crop rows with an average angular error of 2.15° and an average displacement error of 3.65 cm. Our method could settle the visual servoing controller to these average values within 23 frames of execution on average.

## ACKNOWLEDGMENTS

This work was supported by Lincoln Agri-Robotics as part of the Expanding Excellence in England (E3) Programme, Beijing Municipal Natural Science Foundation (Grant No. 4214060), and National Natural Science Foundation of China (Grant No. 62102443).

## DATA AVAILABILITY STATEMENT

The data that support the findings of this study are openly available in CropRow Detection Data set at https://github.com/JunfengGaolab/CropRowDetection.

## ORCID

*Rajitha de Silva*  http://orcid.org/0000-0001-6404-1715
*Grzegorz Cielniak*  http://orcid.org/0000-0002-6299-8465
*Junfeng Gao*  https://orcid.org/0000-0002-5622-8210

## APPENDIX A: CROP ROW DETECTION ERRORS

Crop row detection errors in each class ($\Delta\theta$, average angular error; $\Delta L_{X2}$, average; $L_{X2}$, error; $\epsilon$, overall crop row detection performance; B, baseline).

| Data class | $\Delta\theta_B$ | $\Delta\theta$ | $\Delta L_{X2,B}$ | $\Delta L_{X2}$ | $\epsilon_B(\%)$ | $\epsilon(\%)$ |
|---|---|---|---|---|---|---|
| 1 | 5.28 | 1.67 | 53.01 | 9.91 | 47.00 | 85.93 |
| 2 | 7.76 | 4.44 | 31.44 | 40.05 | 40.46 | 57.22 |
| 3 | 3.50 | 1.25 | 41.59 | 10.93 | 62.31 | 88.09 |
| 4 | 3.98 | 1.97 | 44.78 | 10.86 | 58.14 | 83.76 |
| 5 | 5.44 | 1.21 | 51.66 | 8.33 | 46.58 | 89.37 |
| 6 | 3.78 | 1.35 | 126.74 | 8.65 | 27.04 | 88.39 |
| 7 | 3.78 | 0.90 | 38.42 | 7.28 | 61.85 | 91.67 |
| 8 | 3.21 | 0.64 | 21.02 | 7.96 | 72.22 | 92.97 |
| 9 | 5.01 | 1.5 | 51.51 | 10.94 | 49.27 | 86.56 |
| 10 | 4.33 | 1.05 | 46.57 | 10.89 | 55.35 | 89.33 |
| 11 | 4.79 | 1.25 | 47.71 | 8.83 | 52.09 | 88.89 |
| 12 | 3.20 | 2.18 | 31.08 | 14.36 | 68.28 | 81.07 |
| 13 | 3.81 | 1.04 | 41.67 | 17.33 | 60.40 | 86.85 |
| 14 | 4.46 | 1.96 | 62.39 | 13.60 | 48.29 | 82.74 |
| 15 | 4.78 | 1.74 | 47.53 | 10.57 | 52.18 | 85.23 |
| 16 | 5.68 | 1.25 | 55.93 | 6.20 | 43.43 | 89.99 |

| Data class | $\Delta\theta_B$ | $\Delta\theta$ | $\Delta L_{X2,B}$ | $\Delta L_{X2}$ | $\epsilon_B(\%)$ | $\epsilon(\%)$ | Data class | $\Delta\theta_B$ | $\Delta\theta$ | $\Delta L_{X2,B}$ | $\Delta L_{X2}$ | $\epsilon_B(\%)$ | $\epsilon(\%)$ |
|---|---|---|---|---|---|---|---|---|---|---|---|---|---|
| 17 | 3.79 | 3.29 | 39.31 | 19.43 | 61.45 | 72.35 | 31 | 4.07 | 3.85 | 37.76 | 41.25 | 60.35 | 60.36 |
| 18 | 4.5 | 1.10 | 49.51 | 5.12 | 53.13 | 91.31 | 32 | 3.90 | 1.00 | 45.26 | 6.92 | 58.45 | 91.17 |
| 19 | 5.08 | 1.59 | 67.60 | 9.51 | 42.45 | 86.56 | 33 | 6.96 | 1.97 | 65.93 | 31.76 | 31.72 | 75.48 |
| 20 | 4.22 | 0.84 | 79.64 | 4.87 | 42.93 | 93.0 | 34 | 5.46 | 1.78 | 38.46 | 10.78 | 51.67 | 84.91 |
| 21 | 2.94 | 0.52 | 53.08 | 4.10 | 61.18 | 95.23 | 35 | 4.71 | 2.84 | 53.85 | 10.14 | 50.13 | 78.72 |
| 22 | 5.15 | 2.24 | 45.46 | 14.81 | 50.79 | 80.53 | 36 | 4.01 | 1.04 | 35.44 | 6.78 | 61.69 | 91.03 |
| 23 | 5.67 | 2.33 | 72.98 | 19.78 | 36.79 | 78.07 | 37 | 4.09 | 2.08 | 37.87 | 9.48 | 60.23 | 83.62 |
| 24 | 3.71 | 1.30 | 32.39 | 9.05 | 64.68 | 88.53 | 38 | 4.19 | 0.99 | 50.52 | 7.98 | 54.59 | 90.83 |
| 25 | 3.53 | 1.08 | 25.54 | 7.66 | 68.47 | 90.42 | 39 | 3.59 | 2.14 | 35.99 | 8.24 | 63.99 | 83.78 |
| 26 | 4.27 | 1.56 | 58.25 | 7.73 | 51.08 | 87.45 | 40 | 3.65 | 1.23 | 36.40 | 5.32 | 63.44 | 90.45 |
| 27 | 7.73 | 2.28 | 48.94 | 10.87 | 33.75 | 81.87 | 41 | 3.30 | 0.71 | 87.74 | 5.08 | 45.31 | 93.65 |
| 28 | 3.23 | 0.64 | 62.35 | 4.87 | 55.77 | 94.18 | 42 | 8.23 | 2.68 | 65.95 | 13.69 | 23.98 | 78.29 |
| 29 | 3.65 | 0.82 | 66.76 | 4.54 | 51.46 | 93.21 | 43 | 3.98 | 2.63 | 25.47 | 30.50 | 65.80 | 72.01 |
| 30 | 3.53 | 0.84 | 66.63 | 8.76 | 52.29 | 91.43 | **Average** | **4.51** | **1.65** | **50.65** | **11.99** | **50.65** | **85.27** |

(Continues)